\documentclass[
reprint,
superscriptaddress,
floatfix,
amsmath,
amssymb,
aps,
notitlepage
]{revtex4-1}
\usepackage{placeins}
\usepackage[hidelinks=true]{hyperref}
\usepackage{xcolor}
\usepackage{threeparttable}
\usepackage{comment}
\usepackage{amsmath}
\usepackage{setspace}
\usepackage{caption}
\usepackage{float}
\usepackage{siunitx}
\usepackage[utf8]{inputenc}
\usepackage{subcaption}
\usepackage{graphicx}% Include figure files
\usepackage{dcolumn}% Align table columns on decimal point
\usepackage{bm}% bold math
\definecolor{greenish}{RGB}{108, 200, 105}
\definecolor{reddish}{RGB}{174,12,48}
\definecolor{blueish}{rgb}{0.12, 0.56, 1.0}
\definecolor{magenta}{RGB}{242, 80, 93}
\hypersetup{
  colorlinks   = true,
  urlcolor     = greenish,
  linkcolor    = blueish,
  citecolor    = magenta
}

\makeatletter
\newcommand*{\rom}[1]{\exp\!andafter\@slowromancap\romannumeral #1@}
\makeatother

\def\t{\mathbf{t}}
\def\T{\mathbf{T}}
\def\N{\mathbf{N}}
\def\argmin{\mathrm{argmin}}
\def\grad{\nabla}
\def\r{\mathbf{r}}
\def\e{\mathrm{e}}
\def\d{\mathrm{d}}

\def\Nehat{\hat{\mathbf{N}}_e}
\def\Nwhat{\hat{\mathbf{N}}_w}
\def\Tehat{\hat{\mathbf{T}}_e}
\def\Twhat{\hat{\mathbf{T}}_w}
\def\Ne{\mathbf{N}_e}
\def\Nw{\mathbf{N}_w}
\def\Te{\mathbf{T}_e}
\def\Tw{\mathbf{T}_w}
\def\ex{\text{e}}
\def\L{l}
\def\r{\mathbf{r}}
\def\e{\mathrm{e}}
\def\d{\mathrm{d}}
\def\that{\hat{\mathbf{t}}}
\def\tip{\textrm{tip}}

\def\Rot{\mathcal{R}}
\def\tone{\hat{\mathbf{t}}^{(1)}}
\def\ttwo{\hat{\mathbf{t}}^{(2)}}

\def\Nhat{\hat{\mathbf{N}}}
\def\Nehat{\hat{\mathbf{N}}_e}
\def\Nwhat{\hat{\mathbf{N}}_w}
\def\Tehat{\hat{\mathbf{T}}_e}
\def\Twhat{\hat{\mathbf{T}}_w}
\def\I{\mathcal{I}}
\def\tip{\textrm{tip}}
\def\q{\mathbf{q}}
\def\balpha{\boldsymbol{\alpha}}

\begin{document}
\title{Morphing of and writing with a scissor linkage mechanism}
\author{Mohanraj A}
\affiliation{Department of Applied Mechanics \& Biomedical Engineering, IIT Madras, Chennai, TN 600036.}
\author{S Ganga Prasath}
\email{sgangaprasath@smail.iitm.ac.in}
\affiliation{Department of Applied Mechanics \& Biomedical Engineering, IIT Madras, Chennai, TN 600036.}

\begin{abstract}
Kinematics of mechanisms is intricately coupled to their geometry and their utility often arises out of the ability to perform reproducible motion with fewer actuating degrees of freedom. In this article, we explore the assembly of scissor-units, each made of two rigid linear members connected by a pin joint. The assembly has a single degree of freedom, where actuating any single unit results in a shape change of the entire assembly. We derive expressions for the effective curvature of the unit and the trajectory of the mechanism's tip as a function of the geometric variables which we then use as the basis to program two tasks in the mechanism: shape morphing and writing. By phrasing these tasks as optimization problems and utilizing the differentiable simulation framework, we arrive at solutions that are then tested in table-top experiments. Our results show that the geometry of scissor assemblies can be leveraged for automated navigation and inspection in complex domains, in light of the optimization framework. However, we highlight that the challenges associated with rapid programming and error-free implementation in experiments without feedback still remain.
\end{abstract}

\maketitle
\section{\label{Introduction}Introduction}
\noindent Mechanisms made of rigid linkages occupy a unique space in engineering design. The function performed by these mechanisms is often encoded in their geometry -- examples include cams, cranks, and gear assemblies~\cite{reuleaux1963kinematics, erdman1997mechanism, bryant2011round}. Their prevalence in applications ranging from microscopic sensors to architectural-scale assemblies can be attributed to ($a$) ease of deployment -- often the entire assembly is controlled by a single actuator; ($b$) scalability -- since these assemblies are geometric in nature, they work independently of their size; ($c$) reconfigurability -- mechanisms are designed to be modular and can be reconfigured to suit pertinent requirements~\cite{meloni2021engineering, thomaszewski2014computational, pellegrino2001deployable, overvelde2017rational, becker2023c, ono2024growth, panetta2019x, toyonaga2024additive}.

One of the simplest assemblies of rigid linkages is a scissor linkage mechanism composed of linear members connected by pin joints~\cite{dinevari2021geometric, maden2011review, xu2023interactive, zhang2016designing, zhang2022inverse}. They carry the advantages of traditional mechanisms such as: $(a)$ actuating one end of the mechanism introduces a correlated motion across all units, allowing the entire assembly to be controlled by this simple action; $(b)$ since the members are rigid, they can be scaled to different sizes without loss in their functionality; $(c)$ they are also modular in that different scissor-units can be assembled together (subject to geometric constraints) to produce complex shape transformations. Given such a capability, the underlying geometry of the assembly has yet to be understood in a manner that can be used for algorithmic programming of the desired functions.

\begin{figure*}[t]
    \centering
    \includegraphics[width=\linewidth]{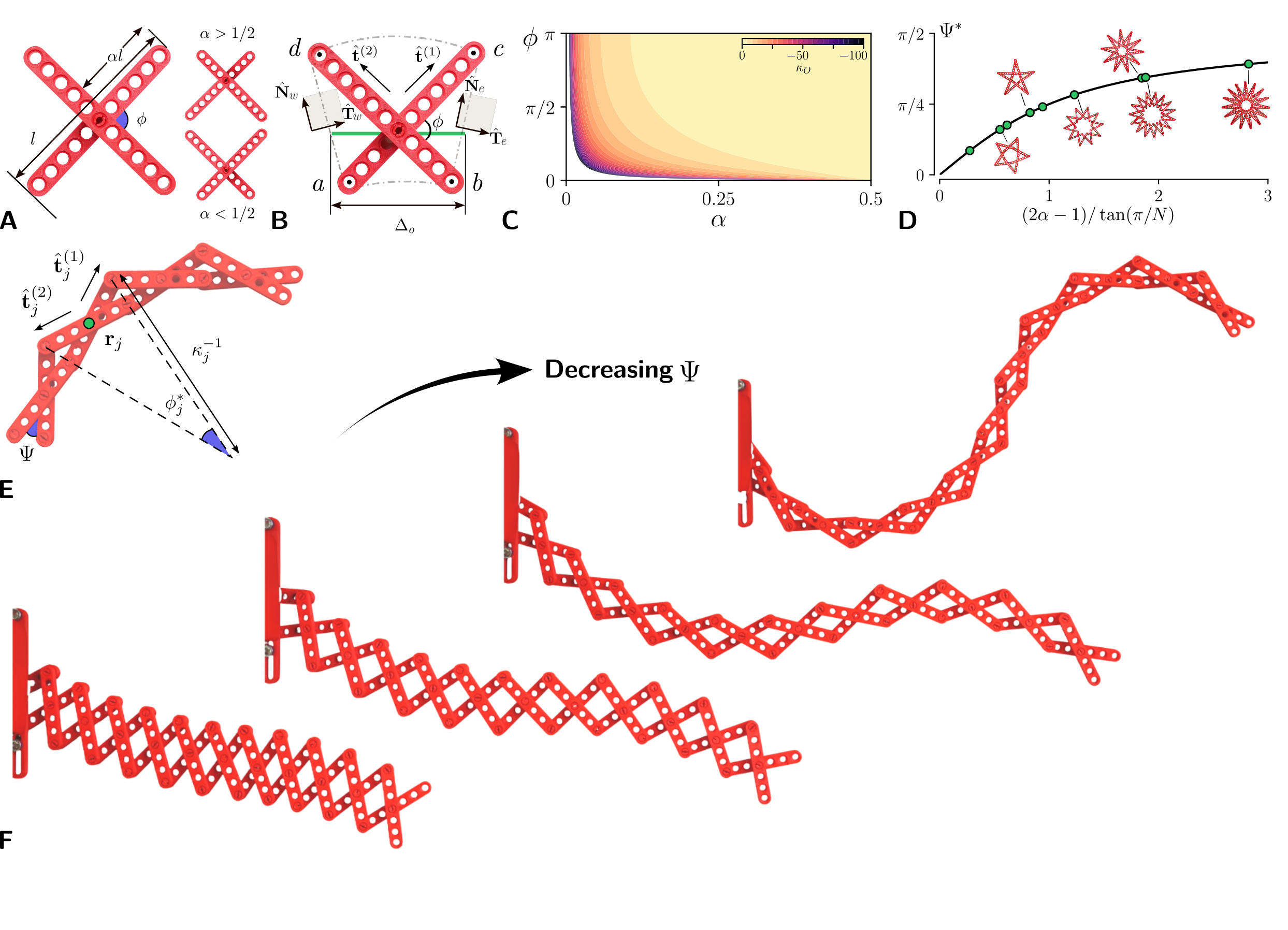}
    \caption{\label{fig:schmUnit} \textbf{\textsf{Geometry of scissor linkage mechanism.}}
    (A) A single scissor-unit is made of two rigid members of length $l$ connected by a pin joint. Each unit is described by two geometric variables: angle between members $\phi$; the aspect ratio $\alpha$ defined as relative distance of the pin joint from the end to the total length, $l$. (B) Vectors $\tone$ and $\ttwo$ capture the orientation of the members $ac$, $bd$. Along the edges $ac$, $bd$ we define two normal vectors $\hat{\mathbf{N}}_e$ and $\hat{\mathbf{N}}_w$. The tangent vectors, $\hat{\mathbf{T}}_e$ and $\hat{\mathbf{T}}_w$, are defined orthogonal to these normal vectors. $\Delta_o$ is the width of the segment passing through the pin joint and intersecting faces $ad, bc$. (C) Contours of $\kappa_o(\alpha, \phi)$, the effective curvature of each scissor-unit defined in Eq.~\ref{eq:kappaFinal}. $\kappa_o$ diverges for $\phi \to 0$ and vanishes at the symmetric midpoint $\alpha=0.5$. (D) Comparison between theoretical prediction of the critical actuation angle $\Psi^*$ (solid line) required to achieve self-intersection ($N\phi^*=2\pi$) and experimentally measured angles (green markers). Error bars are smaller than the size of the markers. (E) Scissor-units are assembled by connecting to neighboring units via pin joints resulting in a single degree-of-freedom for the mechanism. The location of center pin joint of $j$-th unit is denoted by $\r_j$ and the radius of curvature by $\kappa_j^{-1}$. Actuation angle $\Psi$ is the angle between the members of the first unit, $\Psi = \phi_1$ and is used to control the motion of the assembly. $\phi^*$ is defined as the angle between the lines connecting the sides $ad$ and $bc$ in Fig. ~\ref{fig:schmUnit}(B)). (F) Sequence of configurations of a scissor mechanism from experiments whose aspect ratios are given by $\{\alpha_j\} = 0.45$ for $0<j<6$ and $\{\alpha_j\} = 0.55$ for $6 \leq j < 12$. Decreasing the actuation angle $\Psi$ via Arduino controlled motor drive (see Sec.~\ref{Model} for details) steers the mechanism from a near-straight configuration (left) to a curved shape (right), illustrating large-scale shape change enabled by the intrinsic geometry.}
\end{figure*}

In this article, we explore the geometry of a scissor mechanism composed of two-fold symmetric units (also referred to as polar units) that are described by a single geometric variable, the relative location of the pin joint from the ends of members. We show that the geometry of this mechanism can be described by an effective curvature, analogous to discrete curves, which depends on the aspect ratio and the state of actuation of each unit. Although programming functions in mechanisms is often challenging~\cite{thuruthel_soft_2019,Santina_2023,naughton_elastica_2021}, we utilize an optimization framework to explore this in two tasks -- shape morphing and writing. The definition of curvature lends itself to the optimization procedure in which the shape error between a desired shape and the given state of the mechanism is minimized with the aspect ratio of each unit as the tunable parameter. Our experiments, using an assembly of custom 3D printed components that is actuated by a servo, help test the solutions from the theoretical model for shape morphing tasks. Since the location of the distal end of the mechanism is sensitive to tiny changes in the aspect ratio of the units in the interior, we use a differentiable simulation framework to evaluate the gradient in the shape error for the writing task accurately while also leveraging the compositional structure of the mechanism. Our results show a unified approach to leverage the geometry of linkage mechanisms to program functions in a variety of tasks.

\section{Model and Experiments}\label{Model}
\noindent The scissor linkage mechanism we study is an assembly of $N$ scissor-units, each of which is made up of two rigid members of equal and fixed length $l$. The rigid members corresponding to the $j$-th unit are connected by a pin joint located at a distance $\alpha_j l$ from one end of both members (see Fig.~\ref{fig:schmUnit}A). Each unit is then described by the aspect ratio, $\alpha_j \in (0, 1)$ and the angle $\phi_j$ between the members (which is also the only internal degree-of-freedom of the unit). The $j$-th unit is connected to its neighboring units via pin joints which retains a single global degree-of-freedom for the entire mechanism. The mechanism for a fixed $\alpha_j$ for $j = 1 \dots N$ can be deployed by actuating the angle of the first unit, which is denoted as the actuation angle, $\Psi = \phi_1$. The members of the $j$-th unit in the assembly are described by two orientation vectors $\that^{(1)}_j$ and $\that^{(2)}_j$ (see Fig.~\ref{fig:schmUnit}E), aligned along their length with $\phi_j$ being the angle between them. The center of the unit, denoted by $\r_j$, is obtained from the coordinates of the $(j-1)$-th unit via the relation,
\begin{equation}
    \r_j
    =
    \r_{j-1}
    +
    l(\alpha_{j-1} \that^{(1)}_{j-1}
    +
    \alpha_j \, \that^{(2)}_{j}).
    \label{eq:discrete_step}
\end{equation}
Equation~\eqref{eq:discrete_step} describes the connectivity constraint and for a given $\Psi$ the orientations $(\that^{(1)}_j,\that^{(2)}_j)$ of each unit are uniquely determined. Decreasing $\Psi$ drives the mechanism from a fully folded state through a sequence of deployed configurations, thus acting as the actuating degree of freedom in our experiments (shown in Fig.~\ref{fig:schmUnit}F).

\begin{figure*}[t]
    \centering
    \includegraphics[width=\textwidth]{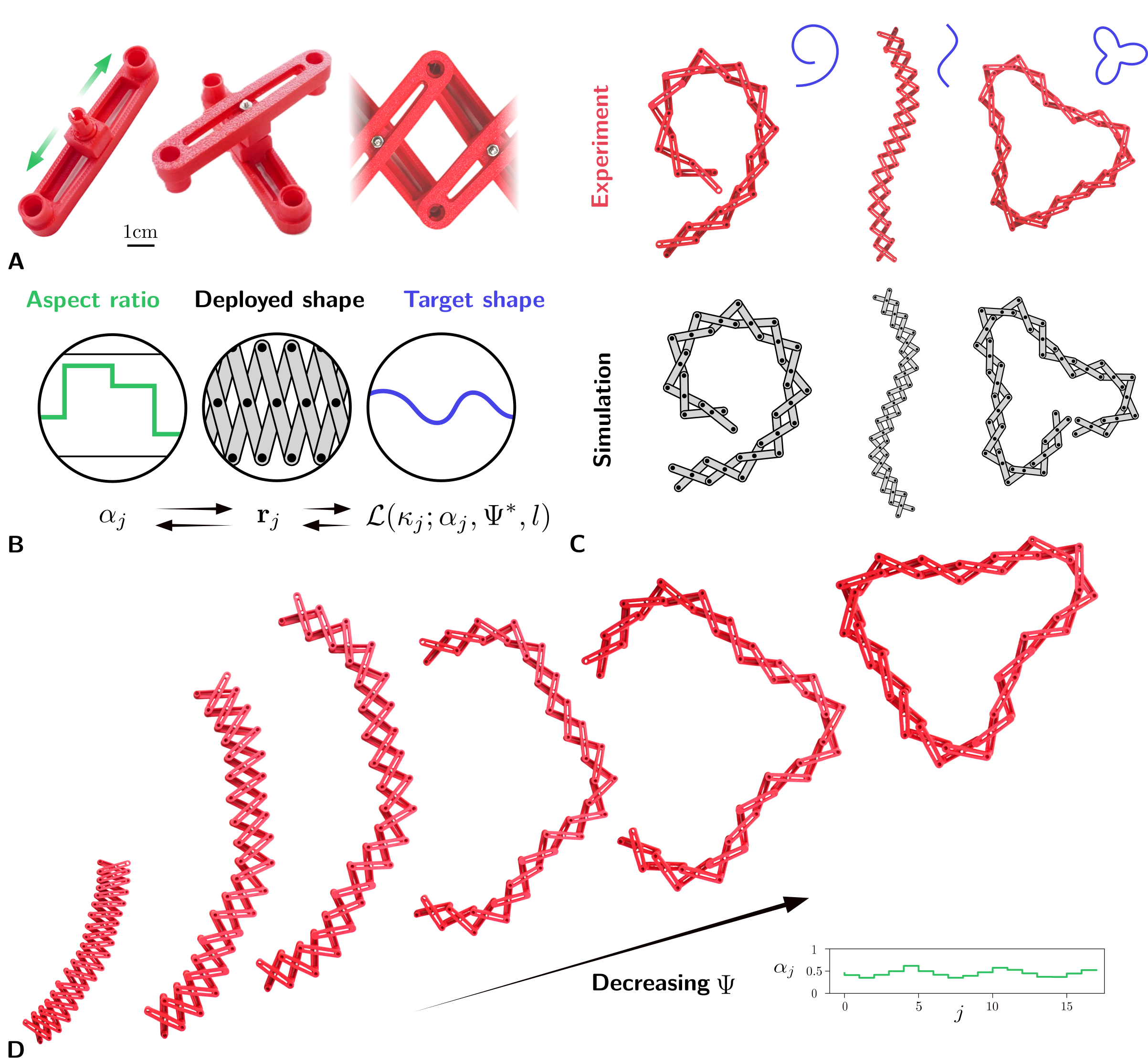}
    \caption{
    \textbf{\textsf{Shape morphing of scissor mechanism.}}
    (A) 3D printed scissor-units made with an adjustable slider allowing continuous variation in the aspect ratio $\alpha$ along with a screw that can be locked into position. These units are connected to neighboring units via pin joints. (B) In the shape morphing task, the mechanism is designed to take a target shape by choosing appropriate aspect ratios $\{\alpha_j\}$ and actuating the mechanism to angle $\Psi^*$. These values are arrived at by minimizing the shape error, $\mathcal{L}$ between the current configuration, $\r_j$ and the target shape described by a curvature $\kappa^t$ (see SI Sec.~\ref{StaticDesignDetails} for details). (C) Examples showing the scissor mechanism deployed to target shapes (curves in blue) in experiments and simulation (see SI Video 2 for the deployment sequence). The target shapes have $(i)$ a monotonically increasing curvature (left); $(ii)$ a sinusoidal curvature profile (middle); $(iii)$ a three-petaled flower (right). (D) Evolution of the mechanism shape in experiments upon actuation by decreasing $\Psi$. Inset shows the solution aspect ratios $\alpha_j$ obtained via the optimization procedure detailed in Sec.~\ref{Static Inverse Design}.}
    \label{fig:invDesign}
\end{figure*}

\subsection*{Experimental Setup}
We demonstrate the capabilities of the scissor mechanism by building a prototype using custom 3D-printed scissor-units, fabricated to ensure dimensional accuracy and to preserve the precise aspect ratios $\alpha_j$ required for shape morphing. The mechanism is assembled on a horizontal planar surface, where its single global degree-of-freedom $\Psi$ is actuated by a TowerPro MG995 Metal Gear Servo (stall torque of 13--15 kg-cm, ref. SI Sec.~\ref{ExpSetup} for details). The motor applies sufficient force to overcome the accumulated friction in the pin joints and the contact friction with the mounting surface. The servo is mechanically coupled to the first unit via a direct-drive transmission that converts the rotary motion of the shaft directly into the relative rotation of the first unit's members (see SI Sec.~\ref{ExpSetup} for a 3D visualization of the assembly). The actuation system is controlled by an Arduino Nano microcontroller, which processes inputs from a linear potentiometer that helps with continuous actuation. The potentiometer allows us to slowly deploy the mechanism and pause at specific angles to compare the static shapes against the model (see SI Sec.~\ref{ExpSetup} for more details).
\begin{figure*}[t]
    \centering
    \includegraphics[width=\textwidth]{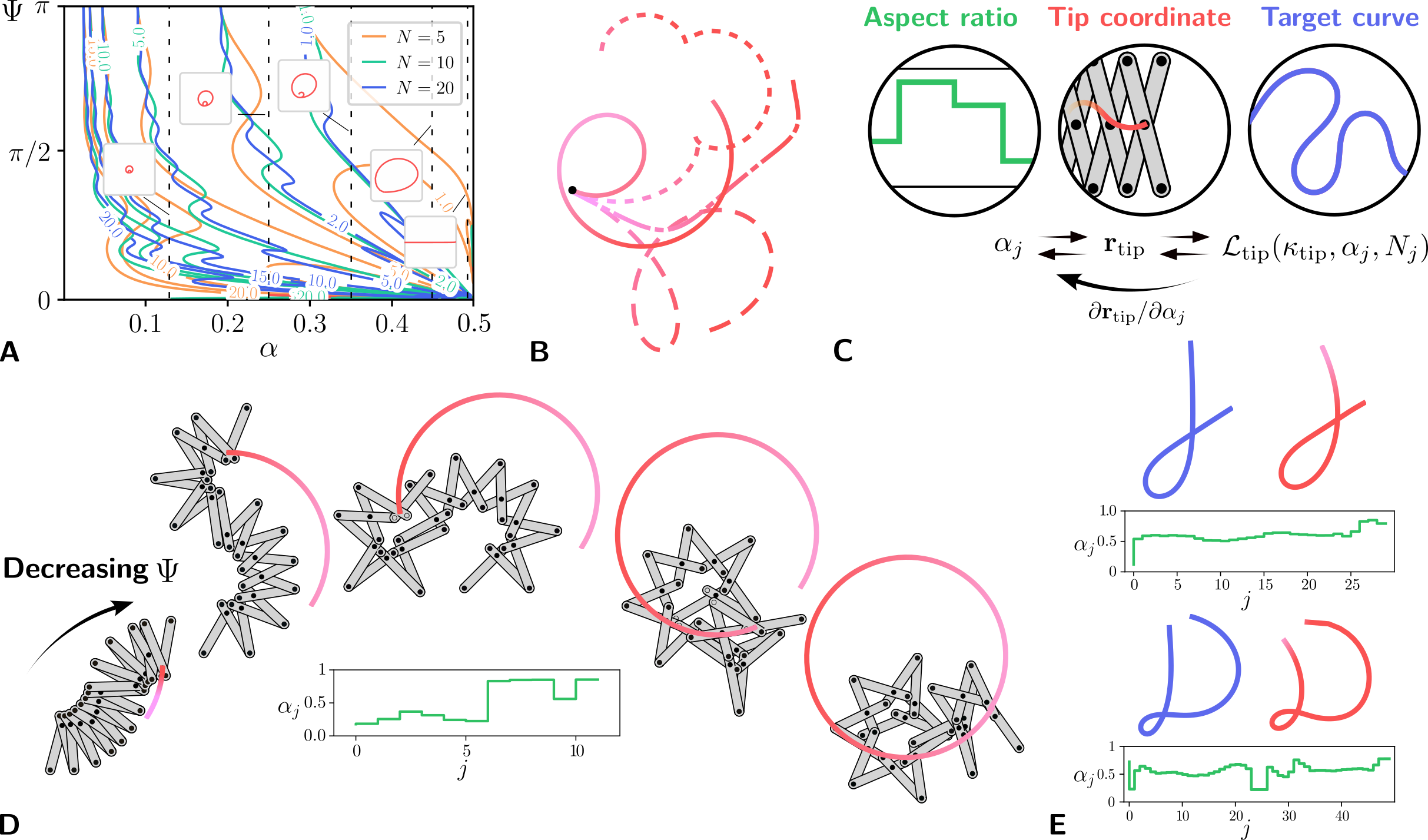}
    \caption{\textsf{\textbf{Writing with a scissor mechanism.}} (A) Contour in the $(\alpha,\Psi)$ plane showing the curvature of the distal end of the mechanism $\kappa_{\tip}(\alpha,\Psi;N)$ made of $N$ scissor-units with constant $\alpha$. Three different contours corresponding to $N=5$ (orange), $N=10$ (green), and $N=20$ (blue) are shown and iso-contour labels correspond to magnitude of $\kappa_{\tip}$. Insets illustrate tip trajectories $\mathbf r_{\tip}(\Psi)$ (for $N=5$) obtained by sweeping $\Psi$ from $\pi$ to $0$ at the corresponding parameter locations. As $\alpha$ departs from $0.5$ the trajectory rapidly develops tight loops (large $\kappa_{\tip}$), whereas $\alpha\to 0.5$ yields an almost straight path with $\kappa_{\tip}\approx 0$. We see that the mechanism is capable of tracing trajectories with rapidly changing tip curvature, however, they vary only monotonically. (B) Representative tip trajectories generated using the forward kinematics formulation for segmented mechanisms in Eq.~\ref{eq:seg_update}. Trajectories are for $J=3$ sections with distinct aspect ratios ($\alpha_j = \{0.41, 0.17, 0.55 \}$ for solid line, $\alpha_j = \{ 0.44,0.43,0.23\}$ for short dashed line, $\alpha_j = \{0.71, 0.32, 0.51 \}$ for long dashed line, $\alpha_j = \{ 0.41, 0.62, 0.89 \}$ for dot-dashed line). The mechanism is capable of tracing complex paths with varying curvature, including spirals and self-intersecting loops. (C)  The mechanism is parameterized by constant aspect ratios $\{\alpha_j\}$ (function in green), which determine the tip trajectory $\r_{\tip}(\Psi)$ (shown in red) through the forward kinematic relations in Eq.~\ref{eq:seg_update}. The resulting trajectory is compared against a prescribed target curve (blue) in the tip-trajectory error $\mathcal L_{\tip}$ (see Eq.~\ref{eq:curv_loss}), which quantifies the curvature mismatch between $\kappa_{\tip}(s;\{\alpha_j,N_j\},\Psi)$ and the target curvature $\kappa^{t}(s)$. The differentiable simulation provides exact derivatives (such as $\partial \mathbf r_{\tip}/\partial \alpha_j$) via automatic differentiation, enabling efficient gradient-based updates of $\{\alpha_j\}$ (for fixed $\{N_j\}$) and the actuation input $\Psi$ range. (D) Snapshots of the scissor mechanism's deployment as the tip traces a circular trajectory. The red colored stroke indicates the tip trajectory, and the inset shows the aspect-ratio, $\{\alpha_j\}$ obtained using the optimization procedure. (E) Comparison between target curves (blue) and tip trajectories (red) when the mechanism traces the characters $j$ and $D$, as a tribute to the automaton artist Jaques-Droz. The color gradient from light to dark indicates tracing direction as $\Psi$ decreases from $\Psi_{\max}$ to $\Psi_{\min}$. Insets show the optimized aspect-ratio profiles $\{\alpha_j\}$, where $j$ indexes the scissor-units. See SI Video 3 for a visualization of the mechanism performing the writing task.}
    \label{fig:kinematics}
\end{figure*}

\section{Results}\label{Results}
\subsection{Curvature of a scissor-unit}
The geometry of a scissor-unit can be described by its \textit{effective curvature}, $\kappa_o$ as a function of $l, \alpha, \phi$ (ref. Fig.~\ref{fig:schmUnit}B). Although there are multiple feasible definitions of the curvature of a scissor-unit (see SI Sec.~\ref{CurvatureDefn} for other definitions), the appropriate definition suitable for our purposes is the change in orientation of the face unit-normals $\hat{\T}_w$ and $\hat{\T}_e$, across the width of the unit, $\Delta_o$. After representing the orientation of the rigid members as $\hat{\t}^{(1)} = \e^{i\beta}$ and $\hat{\t}^{(2)} = \e^{i(\pi-\phi)} \e^{i\beta}$ and a bit of algebra, we obtain the expression for the $\kappa_o(\alpha,\phi,l)$ as (see SI ~\ref{TangentChangeDerivation} for details),
\begin{equation}
    \kappa_o(\alpha,\phi,l)
    =
    \frac{(2\alpha-1)}{2\alpha l (1-\alpha)}
    \,\frac{1}{\sin(\phi/2)},
    \label{eq:kappaFinal}
\end{equation}
where we have used $\Delta_o=4\alpha l (1-\alpha)\cos\!\left({\phi/2}\right)$. Figure~\ref{fig:schmUnit}C shows a visualization of $\kappa_o(\alpha,\phi,l)$ as a function of $(\alpha,\phi)$ and we see that $(i)$ $\kappa_o = 0$ as $\alpha = 0.5$, which implies that as the unit tends toward a 4-fold symmetric state, the action of an assembly of $\alpha_j = 0.5$ will result in straight extension; $(ii)$ when $\phi \to 0$ we get $\kappa_o \to \infty$, implying that the arms approach a folded configuration, producing a sharp change in orientation; $(iii)$ the sign of $\kappa_o$ is determined by $(2\alpha-1)$: $\alpha>0.5$ produces local positive curvature, $\alpha<0.5$ produces local negative curvature.

\subsection{Mechanism with constant \texorpdfstring{$\alpha$}{alpha}}
When a scissor mechanism is composed of units with constant aspect ratio, $\alpha_j=\alpha$, $\kappa_o(\Psi)$ is only a function of the actuation angle, $\Psi$. Such a mechanism naturally evolves towards a circular configuration upon actuation (shown in the inset of Fig.~\ref{fig:schmUnit}D). The orientations of the members of the $j$-th unit can be written as $\that_{j}^{(1)} = \Rot(\phi^*)^{\,j} \,\that_{0}^{(1)}, \that_{j}^{(2)} = \Rot(\phi^*)^{\,j} \,\that_{0}^{(2)},$ with $\that_{0}^{(1)} = \e^{i\beta_0}, \that_{0}^{(2)} = \e^{i(\pi-\phi)}\e^{i\beta_0}$, where $\Rot(\phi^*)$ is the two-dimensional rotation matrix and for a constant $\alpha$, every unit undergoes rotation by an angle $\phi^*$ relative to its neighbor (shown in Fig.~\ref{fig:schmUnit}E). From this we can calculate $\r_j=    \Rot(\phi^*)^{\,j}\,\hat{\r}_0\,\left(\kappa^{-1}\right)$. A circle is formed when the orientation of the final unit coincides with that of the first, i.e., $\that_{N}^{(1)} = \that_{0}^{(1)}$. Using the condition $N\,\phi^* = 2\pi$, we can calculate the actuation angle at intersection, $\Psi^*$ (see SI Sec. ~\ref{ActuationAngle} for details) to be $\Psi^* = 2\,\tan^{-1}\!\big[ (2\alpha - 1)/\tan(\pi/j) \big]$. In Fig.~\ref{fig:schmUnit}D we see a good match between the experimental estimate of $\Psi^*$ and the prediction. Furthermore, changing the aspect ratio of a scissor-unit changes the global shape of the mechanism as we tune the actuation angle $\Psi$ (see Fig.~\ref{fig:schmUnit}F and SI Video~1). Even small changes in the magnitude of $\alpha_j$ have amplified effects on the shape of the actuated state (see SI Sec.~\ref{sec:SIpert} for deviations from circle) and this sensitivity can be used to perform large-scale shape change to desired configurations.

\subsection{Shape morphing of scissor mechanism}\label{Static Inverse Design}
\noindent Shape morphing structures have a variety of use cases in applications that involve high-dexterity tasks in confined spaces (such as inspection of devices and surgical procedures in engineering and medicine ~\cite{7139597,5714753, WebsterIII,Runciman}). The geometry of a scissor mechanism, as we have seen in Eq.~\eqref{eq:kappaFinal} for $\kappa_o(\alpha,\phi,l)$, can be tuned by modifying the aspect ratio, $\alpha$ of each unit (see Fig.~\ref{fig:schmUnit}F for an example). We do this in experiments by designing rigid members with an adjustable slider and locking pin, allowing the $\alpha_j$ to be set independently for each unit (shown in Fig.~\ref{fig:invDesign}A). To identify this $\{\alpha_j\}$ required to morph the mechanism to a specified target shape at a prescribed actuation angle $\Psi^*$, we phrase the task as the minimization of a shape error between a given configuration of the mechanism (described by $\kappa_j(\alpha_j,\phi(\Psi),l)$) and the target shape (described by a curvature, $\kappa_j^t$). The task (shown schematically in Fig.~\ref{fig:invDesign}B) then translates to minimizing the shape error,
\begin{equation}
\mathcal{L}[\{\alpha_j\},\Psi] = \frac{1}{N}\sum_{j=1}^{N}\Big(\kappa_j(\alpha_j,\phi(\Psi),l)-\kappa_j^t\Big)^2.
\label{eq:inv_design_opt}
\end{equation}
The solution,
$(\{\alpha_j^*\},\Psi^*)= \argmin_{\{\alpha_j\},\,\Psi} \mathcal{L}(\{\alpha_j\},\Psi)$, yields the intrinsic aspect ratios that encode $\kappa_j^t$ within the scissor geometry. The results of this procedure are shown in Fig.~\ref{fig:invDesign}C, where the scissor mechanism encoded with $\alpha_j^*$ transforms to targets in both simulation and experiments. Fig.~\ref{fig:invDesign}D shows the deployment of the mechanism for a complex shape from the initial state to the target shape as $\Psi$ decreases (see also SI Video 2).

\subsection{Writing with scissor mechanism}

\noindent An allied task we look at is that of encoding motion of the mechanism's distal end along a prescribed trajectories~\cite{Zhu2012MotionguidedMT,Computationaldesignofmechanicalcharacters}, which we call writing. The challenge of encoding complex trajectories into linkages has a rich history in both the artistic and engineering literature. In the arts, this pursuit flourished during the age of automata, where complex cam-driven machines like the Jaquet-Droz's \textit{Writer} or Maillardet's \textit{Draughtsman} were crafted to mimic human handwriting and drawing with uncanny realism~\cite{riskin2003eighteenth}. Parallel to this, engineers were on the hunt for a mechanism that can execute precise straight-lines~\cite{moon2007machines}. This endeavor culminated in Kempe's seminal work~\cite{kempe1876general}, where he identified a procedure to design a linkage mechanism that can trace any algebraic curve. While automata rely on dedicated hardware and Kempe's constructions are often theoretically complex and impractical to manufacture (as he himself acknowledged), we seek a modular alternative using simple, repeatable scissor-units~\cite{teshigawara2019mobile, liu2007automated}.

Although the scissor mechanism is a single degree-of-freedom system, we have seen that the geometry of the deployed state can be encoded by the appropriate choice of $\{\alpha_j\}$. However, the location of the distal end of the mechanism $\r_{\text{tip}} = \mathbf r_{j=N}$ has a non-local coupling to the scissor-units in the interior of the mechanism and is sensitive to small changes in $\{\alpha_j\}$. Further tracing a given trajectory creates a strong history dependence of the mechanism's coordinates as a function of the actuation. This sensitivity and complexity is taken into account by developing a semi-analytic approach coupled to a differentiable framework to compute the distal end trajectory for the writing task. Such a procedure allows for an accurate, efficient, and gradient-based computation of the properties of the mechanisms to allow the tracing of a target trajectory.\\

\noindent \textit{Forward kinematics of the tip:}
The coordinate of the tip of the mechanism,
$\mathbf r_{\text{tip}}(\Psi)$ varies continuously with the actuation $\Psi$. For a mechanism composed of $N$ scissor-units with constant aspect ratio $\alpha$, each unit undergoes the same rotation $\phi^*(\alpha,\Psi)$. The resulting tip trajectory can be explicitly written as (see SI Sec.~\ref{SI:segFK} for details),
\begin{equation}
\mathbf r_{\text{tip}}(\Psi)
=
\mathbf r_0
+
\kappa^{-1}(\alpha,\Psi)
\left(
\mathcal{R}^{\,N-1}(\phi^*) - \mathcal I
\right)\hat{\mathbf p}_1 ,
\label{eq:tip_single}
\end{equation}
where $\mathbf r_0$ is the base position, $\hat{\mathbf p}_1$ is the initial orientation
of the first unit, and $\mathcal I$ is the identity matrix.
Figure~\ref{fig:kinematics}A shows the curvature of the trajectory traced by the tip when varying $\Psi$ for a mechanism with constant $\alpha$ and different number of units $N$. We see that by choosing appropriate $\alpha$ and $N$, the curvature of the tip trajectory can be tuned, however, limiting the curvature to be positive.

The scissor mechanism can trace more complex trajectories when we assemble $J$ contiguous sections $j=1,\dots,J$, where section $j$ contains $N_j$ units of constant aspect ratio $\alpha_j$. Each section is characterized by a constant angle $\phi_j^*(\alpha_j,\Psi)$ and an associated curvature $\kappa_j(\alpha_j,\Psi)$. The forward kinematics can be written in terms of a section-wise center of curvature $\tilde{\q}_j$ and a local orientation vector $\hat{\mathbf p}_j$ as (detailed derivation is in SI Sec.~\ref{SI:segFK}),
\begin{equation}
\begin{aligned}
\tilde{\mathbf{q}}_1 &= \mathbf r_0 - \kappa_1^{-1}(\alpha_1,\Psi)\,\hat{\mathbf p}_1, \\
\tilde{\mathbf{q}}_{j+1} &= \tilde{\mathbf{q}}_j + {\zeta}_j^{j+1}\,\mathcal R(\Omega_j)\hat{\mathbf p}_j, \\
\hat{\mathbf p}_{j+1} &= \mathcal R(\Omega_j)\hat{\mathbf p}_j, \qquad j=1,\dots,J-1, \\
\r_{\text{tip}}(\Psi) &= \tilde{\mathbf{q}}_J + \kappa_J^{-1}(\alpha_J,\Psi)\,\mathcal R(\Omega_{J})\hat{\mathbf p}_J.
\end{aligned}
\label{eq:seg_update}
\end{equation}
Here $\mathbf r_0$ is the base position (center of the first unit) and $\hat{\mathbf p}_1$ is the initial orientation at the base, $\tilde{\q}_j$ is the center of curvature associated with section $j$, and $\hat{\mathbf p}_j$ is the local orientation vector at the start of section $j$. The term ${\zeta}_j^{j+1}$ is the shift in the center of rotation that enforces continuity at the interface between sections $j$ and $j{+}1$ (see SI Sec.~\ref{SI:segFK} for its explicit form). The rotation angles for each  are $\Omega_{1,J} = {( (2N_{1,J}-1)/2)}\phi_{1,J}^*$ and $\Omega_j  = N_j\phi_j^*$ for $j=2,\dots,J-1$. Equation~\ref{eq:seg_update} provides a simplified representation of the tip trajectory, $\r_{\text{tip}}(\Psi)$ and in Fig.~\ref{fig:kinematics}B we see that the mechanism can trace trajectories with spatially varying curvatures. Using this representation, we now proceed to look at the writing task using the optimization framework.\\

\noindent \textit{Differentiable simulation of writing:}
As we see from Eq.~\ref{eq:seg_update}, making the mechanism trace a specific trajectory is computationally expensive due to the non-local nature of the coupling between the tip position, $\r_\tip(\Psi)$ and the aspect ratios, $\{\alpha_j\}$ in the interior of the mechanism (see SI Sec.~\ref{Sensitivity} for the sensitivity analysis) as well as the history dependence of the coordinates $\r_j$ along $\Psi$. Consequently, to identify the $\{\alpha_j\}$ that makes the mechanism trace a desired trajectory using an optimization procedure similar to Sec.~\ref{Static Inverse Design}, we need an efficient method to evaluate $\r_\tip(\Psi)$ and gradients in the mechanism variables, $\{{\alpha_j}\}$. We develop a differentiable simulation framework that allows for fast and accurate evaluation of the tip position to minimize the tip trajectory error, $\mathcal{L}_\tip(\{\alpha_j, l\})$. The tip of the scissor mechanism is made to trace a desired trajectory, described by the curvature $\kappa^{t}(s)$, with $s$ being the arc-length, by minimizing the error,
\begin{equation}
\mathcal{L}_\tip[\{\alpha_j, l\}]
=
\frac{1}{M}\sum_{i=1}^M
\big[
\kappa_\tip(s_i; \{\alpha_j, l\}, \Psi)
- \kappa^{t}(s_i)
\big]^2.
\label{eq:curv_loss}
\end{equation}
Equation~\eqref{eq:seg_update} is a differentiable expression and can be used to evaluate $\kappa_\tip(\Psi)$ for the actuation sweep $\Psi \in [\Psi_{\max},\Psi_{\min}]$ reparameterized in arc-length, $s$. The design variables are the aspect ratios $\alpha_j$ and the unit length $l$, while the number of units $N_j$ is a fixed integer chosen \textit{a priori} (method used to choose $N_j$ is discussed in the SI Sec.~\ref{sec:GridSearch}). We solve for $\{\alpha_j\}^* = \argmin_{\{\alpha_j\}} \mathcal{L}(\{\alpha_j\})$. We implement the computation of the tip trajectory $\r_\tip(s)$ as a sequence of elementary differentiable operations using automatic differentiation and this provides exact gradients of the tip trajectory error $\mathcal{L}$ with respect to the design variables $(\{\alpha_j\},l)$ at $M$ equally spaced points along $s$. We can write the gradient explicitly as,
\begin{equation}
\grad_{\alpha_j} \mathcal{L}_\tip
=
\sum_{i=1}^M
\frac{2}{M}\big(\kappa_\tip(s_i)-\kappa^{t}(s_i)\big)
\bigg( \frac{\partial \kappa_\tip(s_i)}{\partial \r_\tip(s_i)} \bigg)
\bigg( \frac{\partial \r_\tip(s_i)}{\partial \alpha_j} \bigg).
\label{eq:chainrule}
\end{equation}
 The term $(\partial \r_\tip/\partial\alpha_j)$, which encodes the kinematics of the mechanism, is the simulation derivative and is computed using Eq.~\eqref{eq:curv_loss} (described in detail in SI Sec.~\ref{DynamicOptDetails}). In Fig.~\ref{fig:kinematics}D, we show the resulting $\{\alpha_j\}$ and the actuation trajectory using this procedure for a circular target trajectory (see SI Video 3 for dynamics). As a tribute to the famous automata artist behind the \textit{Writer}, Jaques-Droz, we also design linkages that can trace the letters `$j$' and `$D$' (shown in Fig.~\ref{fig:kinematics}E, also see SI Video 3).

\section{Discussion and conclusion}
\noindent In this article we have explored the geometry of scissor mechanisms and examined their capability in two tasks: shape-morphing and writing. Our results demonstrate that these mechanisms, though simple at first glance, can perform a diverse set of tasks despite possessing only a single degree-of-freedom. This versatility arises out of the aspect ratio that can be tuned at the individual unit level. Although our optimization approach successfully encodes function into the mechanism's geometry, achieving error-free experimental implementation remains an important challenge. Small errors in fixing the aspect ratio of the units in experiments, we have seen, result in large shape errors, owing to complex relationship between the geometric variables. This is a well-known issue in robotic systems that undergo large displacement/deformation, which is addressed in advanced systems by embedding feedback modalities to perform dynamic error correction. A solution for passive systems, like the scissor mechanisms, still remains to be found. 

In our study, we have considered scissor units with pin joint position of each unit being fixed, but dynamically tuning their locations using internal actuators simplifies the computations associated with tracing complex trajectories (example shown in SI Video 4). This insight comes from the fact that the curvature is primarily a function of $\alpha$ and dynamically tuning it bypasses the history dependence of the mechanism's configuration. Although implementing such a mechanism with parallel tunability is experimentally challenging, we show that scissor mechanisms in their current form are sufficient to navigate tortuous environments, common in inspection and robotic surgery.

\smallskip
\noindent \textbf{Acknowledgments}: We thank IIT Madras and ANRF-ECRG/2024/003341/ENS for partial financial support. We also thank the INTERFACE lab members for their interactions. The simulation notebooks are available at \href{https://github.com/sgangaprasath/ScissorMechanism2026/}{Github}.
\bibliographystyle{apsrev4-1}
\bibliography{ScissorMechanism}% Produces the bibliography via BibTeX.

%\sectionfont{\fontsize{15}{15}\selectfont}
%\subsectionfont{\fontsize{15}{15}\selectfont}
%\subsubsectionfont{\fontsize{15}{15}\selectfont}
% \large
\widetext
\clearpage
\onecolumngrid
\begin{center}
\textbf{\large Supplemental Information for \\[.2cm]
``Morphing of and writing with a scissor linkage mechanism''}\\[.2cm]
Mohanraj A$^{1}$, S Ganga Prasath$^{1}$\\[.1cm]
{\small \itshape ${}^1$Department of Applied Mechanics \& Biomedical Engineering, IIT Madras, Chennai, TN 600036.\\
}
\end{center}

\setcounter{equation}{0}
\setcounter{figure}{0}
\setcounter{table}{0}
\setcounter{page}{1}
\setcounter{section}{0}
\makeatletter
% \renewcommand{\theequation}{S\arabic{equation}}
% \renewcommand{\thefigure}{S\arabic{figure}}
% \renewcommand{\bibnumfmt}[1]{[S#1]}
%\renewcommand{\citenumfont}[1]{S#1}
%%%%%%%%% Prefix a "S" to all equations, figures, tables and reset the counter %%%%%%%%%%
%\linespread{1.5}
\setstretch{1.5}

%%%%%%%%% Prefix a "S" to all equations, figures, tables and reset the counter %%%%%%%%%%
%\renewcommand{\bibnumfmt}[1]{[S#1]}
%\renewcommand{\citenumfont}[1]{S#1}
%\renewcommand{\thesection}[1]{S#1\arabic{section}}
\renewcommand{\thetable}{S\arabic{table}}%
\renewcommand{\thesection}{S\arabic{section}}
\renewcommand{\thesubsection}{SS\arabic{subsection}}
\renewcommand{\theequation}{S\arabic{equation}}
\renewcommand{\thefigure}{S\arabic{figure}}
%\renewcommand{\thesubsection}{SS\arabic{subsection}{}}
%\renewcommand{\thesubsection}{\roman{subsection}{}}
%\renewcommand{\thesubsubsection}{(\roman{subsubsection}{})}
%%%%%%%%% Merge with supplemental materials %%%%%%%%%%
%%%%%%%%% Prefix a "S" to all equations, figures, tables and reset the counter %%%%%%%%%%
\section{SI videos}
\noindent \href{https://drive.google.com/file/d/1c2gPJSZhZ40r9K0iOLsEzSGljvwg6G2h/view?usp=share_link}{\textbf{Video 1: Deployment of scissor mechanism.}} Deployment of the scissor mechanism with different aspect ratios, $\alpha$ and number of units, $N$. The video begins with the baseline case of $\alpha = 0.5$ ($N=5$), which deploys into a straight line extension, followed by a general case with a random distribution of $\{\alpha_j\}$ ($N=9$) that results in a complex, non-uniform shape. Subsequently, the video shows deployments for $\alpha = 0.6$ (with $N=5,6,7$) and $\alpha = 0.7$ (with $N=5,7,9$). In addition to these cases, two further videos are provided in which the length of each scissor unit is smaller than that shown earlier. In the first of these, all units have an aspect ratio of $\alpha = 0.53$ with $N = 9$. In the second, the mechanism consists of $N = 9$ units, where the first four units have $\alpha = 0.47$ and the remaining five units have $\alpha = 0.53$.

\noindent \href{https://drive.google.com/file/d/1jgK5u94td0bRxdUh-g3nlOmymhWn_khs/view?usp=share_link}{\textbf{Video 2: Shape morphing of scissor mechanism.}} Experimental validation of the shape morphing task for three distinct target geometries: a spiral with linearly increasing curvature, a sinusoidal curve, and a closed three-petaled flower shape. The visualization tracks the physical deployment of the mechanism, where the internal aspect ratios $\{\alpha_j\}$ have been optimized via Eq.~\ref{eq:inv_design_opt} to match the prescribed target curvature $\kappa^t$. As the actuation angle $\Psi$ is decreased (manually deployed by hand), the mechanism evolves continuously to the target configuration.\\

\noindent \href{https://drive.google.com/file/d/1XynuM8aoFZdowCHdPQWS0Aq1ibBvAmXR/view?usp=share_link}{\textbf{Video 3: Dynamics of trajectory tracing.}} Demonstration of the writing capabilities of the mechanism enabled by dynamic trajectory optimization. The animation displays the mechanism's tip tracing four specific profiles: a circle, the cursive letter `$e$', and the characters `$j$' and `$D$'. As the actuation angle $\Psi$ decreases, the tip of the mechanism follows the path defined by the target curvature $\kappa^{t}(s)$.\\

\noindent \href{https://drive.google.com/file/d/16CPA4_YaxOb31jqjbTD_zO8n6lHQQ7mK/view?usp=share_link}{\textbf{Video 4: Trajectory tracing with dynamic aspect ratio.}} Simulation results for the writing problem when the aspect ratio, $\alpha_j$ is allowed to be a function of the actuation angle, $\Psi$. The animation displays the mechanism tracing two profiles: a circle and a three-petaled flower. The optimization problem gets considerably simplified in this scenario.

\section{Curvature Definitions} \label{CurvatureDefn}

In the main text in Eq.~\ref{eq:kappaFinal}, the local curvature $\kappa_o$ of a scissor-unit is defined via the orientation change of tangents across its geometric width. Here, we briefly show that alternative geometric definitions lead to the same qualitative behavior and also look at some useful limits. These constructions are not used in the forward or inverse problems; they serve only to highlight the non-unique nature of the curvature definition (often seen in discrete differential geometry).

\begin{figure*}[t]
    \centering
    \includegraphics[width=\textwidth]{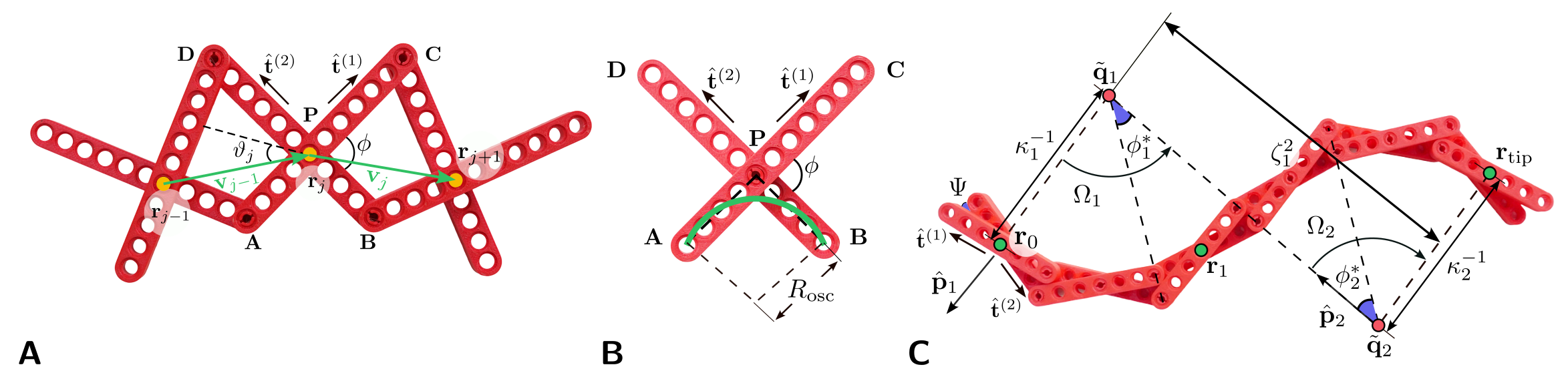}
   \caption{
     (A) Three scissor-units with the same aspect ratio are assembled by connecting neighboring units. Their center nodes are marked by the yellow points $\mathbf{r}_{j-1}$, $\mathbf{r}_{j}$, and $\mathbf{r}_{j+1}$. Using the turning-angle definition, the curvature is given by the angle $\vartheta_j$ between the vectors joining adjacent center nodes, namely $\mathbf{v}_{j-1}$ and $\mathbf{v}_{j}$ (visualized in green). The associated length scale is defined as $|\mathbf{v}_{j-1}|+|\mathbf{v}_{j}|$. (B) For a single scissor-unit, the green circle is the osculating circle of radius $R_{\mathrm{osc}}$; the member orientation vectors $\tone$ and $\ttwo$ are tangent to this circle. (C) A two-segment scissor mechanism for an arbitrary choice of the base center node position $\mathbf{r}_0$ and the initial orientation $\hat{\mathbf{p}}_1$. For a given actuation angle $\Psi$, each section $j$ is modeled as a circular arc with center of curvature $\tilde{\q}_j$ and radius $\kappa_j^{-1}$. At the interface between sections (node $\mathbf{r}_1$), the change in aspect ratio from $\alpha_1$ to $\alpha_{2}$ requires a shift $\zeta_j^{j+1}$ between successive centers of curvature to maintain geometric continuity (from $\tilde{\q}_1$ to $\tilde{\q}_2$ ). The final tip position $\mathbf{r}_{\text{tip}}$ is obtained by iterating these section-wise rotations and translations.}\label{fig:SIfig1}
\end{figure*}

\subsection{Curvature defined by change in tangent orientation}\label{TangentChangeDerivation}
In this section, we derive the expression for the effective curvature $\kappa_o(\alpha,\phi,l)$ presented in the main text. As defined in the main text, the curvature is quantified by the change in orientation of the unit face tangents, $\Tehat$ and $\Twhat$, across the width of the unit, $\Delta_o$. This is analogous to the definition of curvature for a 2D curve, 
\[
    \frac{\d \hat{\mathbf{t}}(s)}{\d s} = \lim_{\Delta s \rightarrow 0 } \frac{\hat{\mathbf{t}}(s+\Delta s) - \hat{\mathbf{t}}(s)}{\Delta_s} = \kappa(s) \mathbf{\hat{n}}(s).
\]
This discrete curvature is defined analogously to the continuous case,
\begin{equation}
    \frac{\Tehat - \Twhat}{\Delta_o} = \kappa_o \N_{\text{avg}}, \quad \text{where} \quad \N_{\text{avg}} = \frac{\Nehat + \Nwhat}{2}.
    \label{eq:curvature_def}
\end{equation}
\noindent The factor of $1/2$ in the definition $\N_{\text{avg}}=(\Nehat+\Nwhat)/2$ is a convention chosen so that $\N_{\text{avg}}$ is an average of the adjacent normals. Any constant rescaling of $\N_{\text{avg}}$ (or, equivalently, of $\Delta_o$) would simply rescale the numerical value of $\kappa_o$ and does not change the qualitative dependence of curvature on $(\alpha,\phi)$ that we use in the main text.
Here, $\Tehat$ and $\Twhat$ are the tangent vectors obtained by rotating the normals by $-\pi/2$. We begin by defining the orientation of the scissor arms in the complex plane as $\tone = \ex^{i\beta}$ and $\ttwo = \ex^{i(\beta + \pi - \phi)} = -\ex^{i(\beta - \phi)}$. The face normals are linear combinations of the orientation vectors along each arm,
\begin{align}
    \Ne &= \alpha \L \tone + (1-\alpha) \L \ttwo, \\
    \Nw &= (1-\alpha) \L \tone + \alpha \L \ttwo.
\end{align}
To utilize Eq.~\eqref{eq:curvature_def}, we compute the difference and the average of these normal vectors. Substituting $\ttwo = -\ex^{-i\phi}\tone$, the difference becomes $\Ne - \Nw = (2\alpha - 1)\L \tone (1 + \ex^{-i\phi})$. Similarly, the average normal vector is found to be $\Nhat_{\text{avg}} = \frac{1}{2}\L \tone (1 - \ex^{-i\phi})$. Substituting these expressions into Eq.~\eqref{eq:curvature_def} and accounting for the rotation of the tangents ($\Te - \Tw = \ex^{-i\pi/2}(\Ne - \Nw)$), we obtain the relation,
\begin{equation}
    (2\alpha - 1) \ex^{-i\pi/2}(1 + \ex^{-i\phi}) \tone = \frac{\L}{2} \kappa_o \Delta_o (1 - \ex^{-i\phi}) \tone .
\end{equation}
Taking the magnitude of both sides allows us to solve for $\kappa_o$. The unit vectors and rotation factors have unit magnitude which implies,
\begin{equation}
    |2\alpha - 1| \, |1 + \ex^{-i\phi}| = \frac{\kappa_o \Delta_o}{2} \, |1 - \ex^{-i\phi}|.
\end{equation}
Using the half-angle identities $|1+\ex^{-i\phi}| = 2\cos(\phi/2)$ and $|1-\ex^{-i\phi}| = 2\sin(\phi/2)$, we solve for $\kappa_o$ to get
\begin{equation}
    \kappa_o(\alpha, \phi, l) = \frac{2(2\alpha-1)}{4\alpha \L (1-\alpha)} \frac{1}{\sin(\phi/2)},
\end{equation}
where we have substituted the expression for the width $\Delta_o = 4\alpha \L (1-\alpha) \cos(\phi/2)$.

\subsection{Curvature defined by neighboring units}
Figure~\ref{fig:SIfig1}A illustrates this construction for a representative configuration of the scissor mechanism, where the discrete centerline is formed by the unit centers, and the turning angle between successive chord vectors defines a local curvature. An alternative description treats the mechanism as a discrete curve formed by the sequence of unit centers $\{\r_j\}$. The turning of this curve provides a curvature measure based on center-to-center geometry. For general (non-uniform) aspect ratios, define $\mathbf{v}_j=\r_{j+1}-\r_j$ and the local turning angle
\begin{equation}
    \vartheta_j = \arg(\mathbf{v}_j)-\arg(\mathbf{v}_{j-1}), \qquad j=1,\dots,N-1.
\end{equation}
We then define the discrete turning curvature at node $j$ in its simplest form as
\begin{equation}
    \kappa_{t,j}=\frac{\vartheta_j}{\Delta_{t,j}},\qquad 
    \Delta_{t,j}=\frac{\|\mathbf{v}_{j-1}\|+\|\mathbf{v}_j\|}{2}.
\end{equation}
This definition reduces to the uniform-unit case when all $\alpha_j$ are identical and $\|\mathbf v_{j-1}\|=\|\mathbf v_j\|$.

For identical units (constant $\alpha$), the geometric compatibility relations of the scissor linkage yield the closed-form expression
\begin{equation}
    \kappa_t(\alpha,\phi,l)
    =
    \frac{
    2\pi -
    \left[
    \phi +
    2\arctan\!\Big((2\alpha-1)\tan\!\big(\tfrac{\pi-\phi}{2}\big)\Big)
    \right]
    }{
    4\alpha l \cos\!\left({\eta(\alpha,\phi)}/{2}\right)
    },
\end{equation}
with $\eta = \pi - \big[2\arctan\!\big((2\alpha-1)\tan ({(\pi-\phi)}/{2})\big) + \phi\big]$.

\noindent \textit{Limiting behavior:} As $\alpha \to 0.5$, the turning angle $\vartheta$ vanishes and $\kappa_t \to 0$, consistent with the symmetric, straight configuration. For $\phi \to 0$, $\kappa_t$ diverges, reflecting the sharp turning of nearly folded units. Quantitatively, $\kappa_t$ differs from $\kappa_o$ by normalization, but both capture the same intrinsic geometric bias introduced by $\alpha \neq 0.5$.

\subsection{Curvature defined by osculating circle}\label{OsculatingCircle}
\noindent The osculating-circle construction is shown in Fig.~\ref{fig:SIfig1}B. Consider the circle tangent to the two members of a single scissor-unit and if the arms' directions subtend an angle $\phi$, trigonometry gives the radius of the osculating circle as
\begin{equation}
    R_{\mathrm{osc}} = (1-\alpha)l \cot\!\left(\frac{\phi}{2}\right),
\end{equation}
with associated curvature $\kappa_{\mathrm{osc}} = 1/R_{\mathrm{osc}}$. This definition depends only on the local arm geometry and again predicts zero curvature in the symmetric limit $\alpha \to 0.5$ and divergence as $\phi \to 0$. While numerically distinct from $\kappa_o$, it encodes the same qualitative geometric information.

\noindent \textit{Remark:} All three definitions $\kappa_o$, $\kappa_t$, and $\kappa_{\mathrm{osc}}$ agree on the sign, symmetry, and singular limits of curvature. However, we adopt $\kappa_o$ in the main text for its direct connection to the rotation of the tangent vector across the unit width (similar to discrete curves in 2D), and its definition depends only on the individual parameters of the scissor-unit.

\section{Actuating towards circular shape}\label{ActuationAngle}

In this section, we consider a \emph{single} scissor-unit (see Fig.~\ref{fig:schmUnit}A for the geometric definitions) and derive the relationship between the unit parameters $(\alpha,\phi)$ and the angle $\phi^*$ (see Fig.~\ref{fig:schmUnit}E) by analyzing the orientation of the unit face normals. Here $\phi$ is the internal angle between the two arms (i.e., the angle between their orientation vectors), and we place the unit symmetrically about the vertical axis so that each arm makes an angle $\phi/2$ with the horizontal.
\medskip

With this convention, the unit vectors representing the orientation of the right and left scissor arms, $\tone$ and $\ttwo$, are
\begin{align}
    \tone &= (\cos(\phi/2), \sin(\phi/2)), \\
    \ttwo &= (-\cos(\phi/2), \sin(\phi/2)).
\end{align}
The face normal vector $\Ne$ is defined as a linear combination of these arm vectors, $\Ne = \alpha \L \tone + (1-\alpha) \L \ttwo$. Substituting the vector components, we calculate the horizontal ($x$) and vertical ($y$) components of $\Ne$,
\begin{align}
    \mathbf{N}_{ex} &= \L \left[ \alpha \cos(\phi/2) - (1-\alpha)\cos(\phi/2) \right] = (2\alpha - 1)\L \cos(\phi/2), \\
    \mathbf{N}_{ey} &= \L \left[ \alpha \sin(\phi/2) + (1-\alpha)\sin(\phi/2) \right] = \L \sin(\phi/2).
\end{align}
Let $\psi$ denote the angle that the normal vector $\Ne$ makes with the horizontal axis. The tangent of this orientation angle is the ratio of the components,
\begin{equation}
    \tan(\psi) = \frac{\mathbf{N}_{ey}}{\mathbf{N}_{ex}} = \frac{\L \sin(\phi/2)}{(2\alpha - 1)\L \cos(\phi/2)} = \frac{1}{2\alpha - 1} \tan\left(\frac{\phi}{2}\right).
\end{equation}
Due to the symmetry of the unit across the vertical axis, the opposing normal $\Nw$ is oriented at an angle $\pi - \psi$. The effective rotation angle $\phi^*$ corresponds to the angle between these two face normals,$\phi^* = (\pi - \psi) - \psi = \pi - 2\psi$. To simplify the final expression, we analyze the half-angle ${\phi^*}/{2} = {\pi}/{2} - \psi$. Taking the tangent of both sides and utilizing the identity $\tan(\pi/2 - \psi) = \cot(\psi)$ yields the relationship between the unit rotation and the standard actuation parameters as,
\begin{equation}
    \tan\left(\frac{\phi^*}{2}\right) = (2\alpha - 1) \cot\left(\frac{\phi}{2}\right) = (2\alpha - 1) \tan\left(\frac{\pi - \phi}{2}\right).
\end{equation}

\section{Effects of perturbation in Aspect Ratio}\label{sec:SIpert}

The main text (see Sec.~\ref{Model} and Fig.~\ref{fig:schmUnit}) focuses on mechanisms with a uniform aspect ratio $\alpha$, for which the deployed configuration is well-approximated by a circle. Here, we study how this picture is modified when $\alpha$ varies slowly along the mechanism. We denote the $j$-th scissor-unit by $\mathcal{S}^j$ and the full mechanism (an assembly of $N$ units) by $\mathcal{S}^m$. We consider $N$ scissor-units indexed by $j=1,\dots,N$, and prescribe a linearly varying aspect ratio for the $j$-th unit, $\alpha_j = \alpha_0 + \epsilon j,$ with $\epsilon \ll 1$. Our goal is to obtain a first-order description of the induced shape by deriving perturbative expressions for the unit configuration and orientation as functions of the slowly varying aspect ratio, and then using these to compute the nodal positions $\r_j$ (and hence $\r_{\text{tip}}$).

\begin{figure*}[t]
    \centering
    \includegraphics[width=0.7\textwidth]{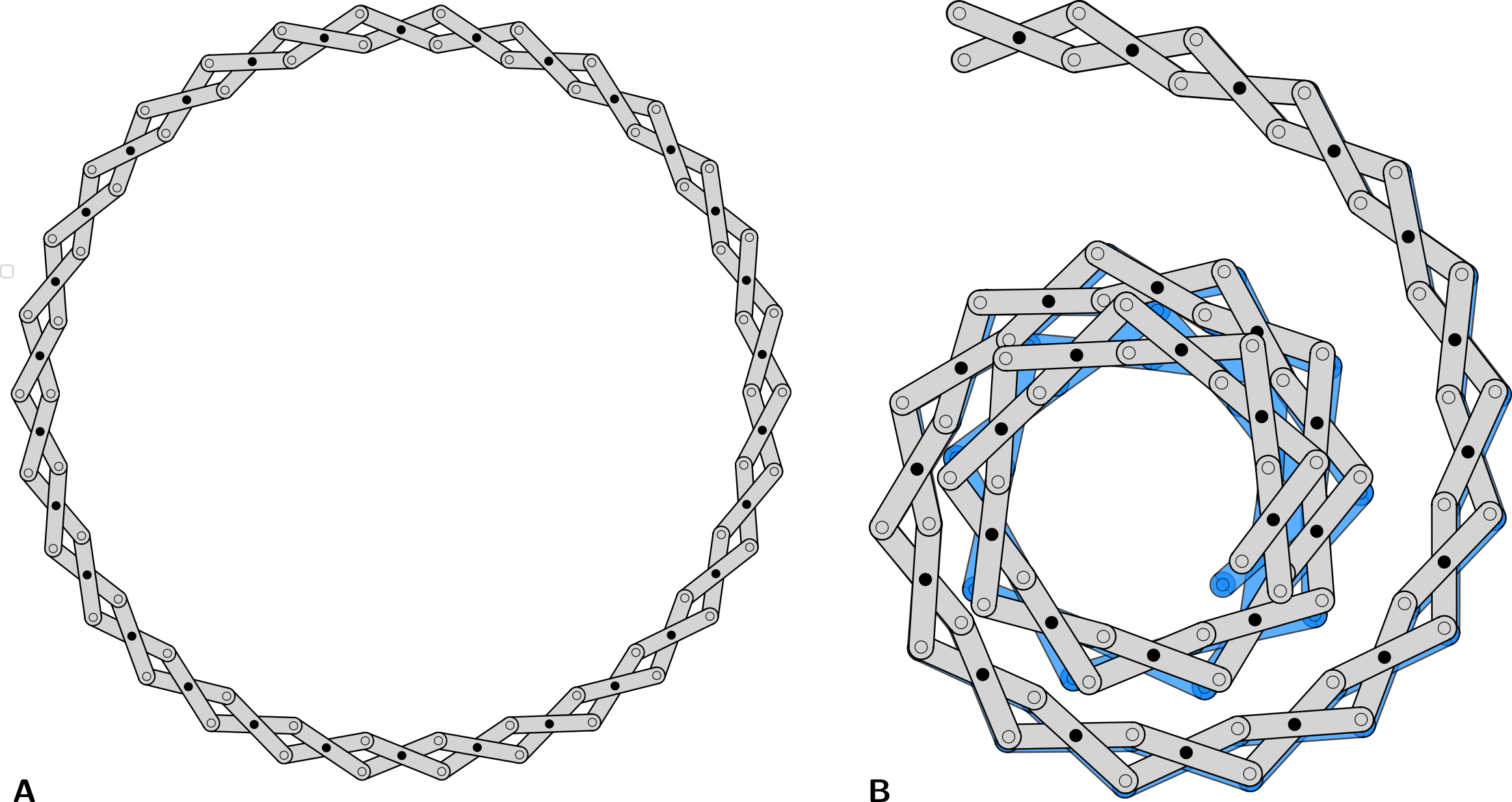}
    \caption{(A) Scissor mechanism corresponding to $N=30$ with $\alpha = 0.52$ and $\Psi = \pi/4 $. (B) Comparison between shape of scissor mechanism with $N=30$ and $\Psi= \pi/4$ constructed from full non-linear simulation (shown in blue) for $\alpha_j = \alpha_0 + \epsilon j$ and perturbative solution in Eq.~\eqref{eq:phi_star_linear} (shown in gray). We see a good match between the perturbative solution and the full solution. We have chosen $\alpha_0 = 0.52, \epsilon = 0.001$.}
    \label{fig:SIfig2}
\end{figure*}

\subsection{Recursion relation for internal angle $\phi_j$}
Consider two neighboring units $\mathcal{S}^j$ and $\mathcal{S}^{j+1}$. The pin-joint constraints form a quadrilateral between the two units; consequently, the internal angle $\phi_{j+1}$ can be expressed as a function of $\alpha_j$, $\alpha_{j+1}$, and $\phi_j$,
\begin{equation}
\phi_{j+1}(\alpha_j,\alpha_{j+1},\phi_j) = \arccos\!\left( \frac{\alpha_{j+1}^2 + (1-\alpha_{j+1})^2 - d(\alpha_j,\phi_j)}{2\,\alpha_{j+1}(1-\alpha_{j+1})} \right),
\end{equation}
where $d$ denotes the diagonal length of the interlocking quadrilateral. For the preceding unit $\mathcal{S}^j$, the law of cosines gives
\begin{equation}
d(\alpha_j,\phi_j)=\alpha_j^2+(1-\alpha_j)^2-2\alpha_j(1-\alpha_j)\cos(\phi_j).
\end{equation}
\noindent Substituting $\alpha_j = \alpha_0 + \epsilon j$ and $\alpha_{j+1} = \alpha_0 + \epsilon(j+1)$ yields
\begin{equation}
\phi_{j+1} = \arccos\!\left( \frac{(\alpha_0+\epsilon(j+1))^2 + (1-(\alpha_0+\epsilon(j+1)))^2 - d(\alpha_0+\epsilon j,\phi_j)}{2(\alpha_0+\epsilon(j+1))(1-(\alpha_0+\epsilon(j+1)))} \right).
\end{equation}
Expanding $\phi_{j+1}$ in $\epsilon$ about $\epsilon=0$ and retaining $\mathcal{O}(\epsilon)$ terms produces the linearized recursion
\begin{equation}
\phi_{j+1} = \phi_j + \epsilon\,\lambda\,\cot\!\left(\frac{\phi_j}{2}\right) + \mathcal{O}(\epsilon^2), \qquad \text{where} \quad \lambda := \frac{2\alpha_0-1}{\alpha_0(\alpha_0-1)}.
\end{equation}

\subsection{Solution to recursion}

Let the initial actuation angle at the base be $\phi_0 = \Psi$. Because the increment in the recursion relation is already of order $\mathcal{O}(\epsilon)$, we can evaluate the cotangent term at the leading-order value $\Psi$,
\begin{equation}
\phi_{j+1} - \phi_j = \epsilon\,\lambda\,\cot\!\left(\frac{\Psi}{2}\right) + \mathcal{O}(\epsilon^2).
\end{equation}
Summing from $k=0$ to $j-1$ then gives
\begin{equation}
\phi_j = \Psi + j\,\epsilon\,\Lambda + \mathcal{O}(\epsilon^2),
\end{equation}
where we have defined the first-order perturbation coefficient $\Lambda := \lambda \cot\left(\Psi/2\right)$.

\subsection{Perturbation expansion of angle $\phi_j^*$}

The quantity $\phi_j^\ast$ (see Sec.~\ref{ActuationAngle}) relates $\phi_j$ and $\alpha_j$ via
\begin{equation}
\phi_j^* = 2\tan^{-1}\!\left[ (2\alpha_j-1)\tan\!\left(\frac{\pi-\phi_j}{2}\right) \right].
\end{equation}
Substituting $\alpha_j = \alpha_0 + \epsilon j$ and the expansion for $\phi_j$, and then expanding consistently to $\mathcal{O}(\epsilon)$, shows that $\phi_j^\ast$ also varies linearly along the mechanism,
\begin{equation}
\phi_j^* = \phi_0^* + \epsilon j\,\mu + \mathcal{O}(\epsilon^2).
\end{equation}
Here, $\phi_0^*$ is the unperturbed turning angle given by $\phi_0^* = 2\tan^{-1}\!\left[ (2\alpha_0-1)\tan\!\left(\frac{\pi-\Psi}{2}\right) \right]$, and $\mu$ is,
\begin{equation}
\mu = \frac{2}{1+(2\alpha_0-1)^2\tan^2\!\left(\dfrac{\pi-\Psi}{2}\right)} \left[ 2\tan\!\left(\frac{\pi-\Psi}{2}\right) - (2\alpha_0-1)\frac{\Lambda}{2} \sec^2\!\left(\frac{\pi-\Psi}{2}\right) \right].
\end{equation}

\subsection{Absolute orientations $\Theta_j^{(m)}$}
To obtain the deployed shape, we compute the absolute orientations $\Theta_j^{(m)}$ of the two members ($m\in\{1,2\}$) with respect to a first unit. For $j\ge 1$, $\Theta_j^{(m)}$ is obtained by summing the contributions from the preceding units,
\begin{equation}
\Theta_j^{(m)} = - \left( \sum_{k=1}^{j-1}\phi_k^\ast + \frac{1}{2}\left(\phi_j^\ast + (-1)^m\phi_j + \phi_0^\ast\right) \right).
\end{equation}
Using $\phi_k^\ast = \phi_0^\ast + \epsilon k\,\mu$ and $\sum_{k=1}^{j-1} k = j(j-1)/2$, we obtain
\begin{equation}
\sum_{k=1}^{j-1} \phi_k^* = (j-1)\phi_0^* + \epsilon \mu \frac{j(j-1)}{2} + \mathcal{O}(\epsilon^2).
\end{equation}
The local contribution from the $j$-th unit follows from the linearized forms for $\phi_j$ and $\phi_j^\ast$,
\begin{equation}
\frac{1}{2}\left(\phi_j^\ast + (-1)^m\phi_j + \phi_0^\ast\right) = \phi_0^* + (-1)^m \frac{\Psi}{2} + \frac{\epsilon j}{2} \left[ \mu + (-1)^m \Lambda \right] + \mathcal{O}(\epsilon^2).
\end{equation}
Substituting and collecting $\mathcal{O}(1)$ and $\mathcal{O}(\epsilon)$ terms yields
\begin{equation}
\Theta_j^{(m)} = - \underbrace{\left[ j\phi_0^* + (-1)^m \frac{\Psi}{2} \right]}_{\text{Leading Order, } \mathcal{O}(1)} - \epsilon \underbrace{\left[ \frac{\mu}{2} j^2 + (-1)^m \frac{\Lambda}{2} j \right]}_{\text{First Order, } \mathcal{O}(\epsilon)} + \mathcal{O}(\epsilon^2).
\end{equation}
Notice that the terms linear in $j$ associated with $\mu$ cancel, leaving a clean explicit first-order expression for the orientation angle. The shape of the scissor mechanism using the perturbed solution is shown in Fig.~\ref{fig:SIfig2} and compares well with the full solution.

\subsection{Orientation vectors and distal tip position}
Adopting the complex-plane representation (planar unit vectors written as $\exp(i\Theta)$), the absolute orientation vectors are obtained by exponentiating $\Theta_j^{(m)}$,
\begin{equation}
\that_j^{(m)} = \exp\left\{ -i \left[ j\phi_0^* + (-1)^m \frac{\Psi}{2} + \epsilon \left( \frac{\mu}{2} j^2 + (-1)^m \frac{\Lambda}{2} j \right) \right] \right\}.
\end{equation}

The position of the distal tip of the mechanism, $\r_{\text{tip}}$, is obtained by the same Forward kinematics expression used in Sec.~\ref{Model}, i.e., by summing the oriented member segments along the mechanism from the first unit to the tip. Substituting the perturbed aspect ratio $\alpha_j = \alpha_0 + \epsilon j$ and our explicit orientation vectors into this accumulation yields,
\begin{equation}
\r_{\text{tip}}(\Psi) = l\left[ \alpha_0 \that^{(1)}_0 + \alpha_1 \that^{(2)}_1 + \sum_{j=2}^{N-1} \left( (\alpha_0 + \epsilon(j-1)) \that^{(1)}_{j-1} + (\alpha_0 + \epsilon j) \that^{(2)}_j \right) \right].
\end{equation}\label{eq:phi_star_linear}
 \label{StaticDesignDetails}
 
 \section{Numerical Implementation of Static Inverse Design}\label{StaticDesignDetails} 
 
This section provides numerical details supporting the static inverse-design formulation introduced in Sec ~\ref{Static Inverse Design} of the main text (Eqn \eqref{eq:inv_design_opt}). The goal is to determine a sequence of aspect ratios $\{\alpha_j\}$ and a single actuation input $\Psi^*$ such that the deployed configuration matches a prescribed target shape at that actuation state.

\begin{figure*}[t]
    \centering
    \includegraphics[width=0.7\textwidth]{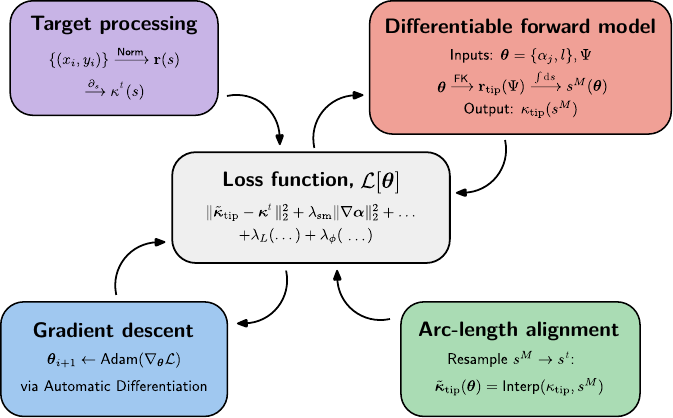}
    \caption{Algorithm used for the differentiable framework for the writing task explained in Sec.~\ref{DynamicOptDetails}. The four steps involved in this framework are: $(i)$ \textbf{Target processing} - normalization of raw target coordinates to a unit bounding box followed by arc-length parameterization to extract the smoothed target curvature profile $\kappa^t(s)$; 
    $(ii)$ \textbf{Differentiable forward model} - evaluation of tip trajectory via forward kinematics and subsequent reparameterization from $\Psi$ to arc length $s$ to obtain the intrinsic mechanism curvature $\kappa_{\text{tip}}(s^M)$ as a differentiable function of arc length; $(iii)$ \textbf{Gradient descent} - iterative optimization of design parameters $\boldsymbol{\theta}$ using the Adam optimizer, with exact gradients of the composite loss $\mathcal{L}[\boldsymbol{\theta}]$ computed via automatic differentiation; $(iv)$ \textbf{Arc-length alignment} - differentiable linear interpolation of the mechanism curvature onto the fixed target arc-length grid to enables a pointwise error comparison.}
    \label{fig:SIfig3}
\end{figure*}

\subsubsection{Discretizing target curve}
To evaluate the target curvature, we discretize the target curve into $M$ segments of equal arc length, yielding $M+1$ nodes corresponding to the boundaries of the $M$ units of the mechanism. For a continuously defined target curve $y=f(x)$, the total arc length $L_{\mathrm{target}}$ is calculated as
\begin{equation}
L_{\mathrm{target}} = \int_0^{x_{\mathrm{end}}} \sqrt{1+\left(\frac{\d y}{\d x}\right)^2}\, \d x.
\end{equation}
The total arc length is divided into $M$ uniform segments $\Delta s = L_{\mathrm{target}}/M$. Rather than implicitly solving the integral constraint, the discrete target node positions $\mathbf{p}_j=(x_j,y_j)$ are obtained by constructing cubic spline interpolants $x(s)$ and $y(s)$ from the densely sampled target curve. The coordinates are then evaluated directly at the uniform arc-length intervals,
\begin{equation}
\mathbf{p}_j = \Big(x(j\,\Delta s),\, y(j\,\Delta s)\Big),
\qquad j=0,\dots,M.
\end{equation}

\subsubsection{Evaluating target curvature}
The curvature values $\{\kappa_j^t\}$ used in the cost function are evaluated at the discretized target nodes. For analytically defined curves $y=f(x)$, the target curvature at node $j$ is computed directly as,
\begin{equation}
\kappa_j^t = \frac{y''}{\left(1+(y')^2\right)^{3/2}}\Bigg|_{x=x_j}.
\end{equation}

\subsection{Discrete forward kinematics}
For a given parameter set $(\Psi,l,\beta_0,\{\alpha_j\})$, the deployed configuration $\{\r_j\}$ is obtained by iteratively enforcing the scissor linkage constraints described in Sec.~\ref{Model} of the main text. Starting from a base unit at the origin, each subsequent unit position is resolved from rigid link-length constraints. This forward kinematics is denoted by $\mathcal{F}:(\Psi,l,\beta_0,\{\alpha_j\}) \longrightarrow \{\r_j\}$. The corresponding discrete curvature values $\kappa_j(\alpha_j,\phi_j,l)$ follow directly from Eq.~\eqref{eq:kappaFinal} in the main text.

\subsubsection{Arc-length reparameterization of mechanism curvature}
The configuration of the mechanism is governed by the global actuation angle $\Psi$ and the geometric parameters $l$ and $\{\alpha_j\}$. Consequently, the local curvature of the $j$-th scissor-unit is a function of these parameters, which we denote by $\kappa_j(\alpha_j,\phi_j,l)$. In contrast, the target curve is defined in Cartesian space and is naturally parameterized by its arc length $s$, with an intrinsic curvature profile $\kappa^t(s)$. 

To define a meaningful point-wise error metric, we must bridge these two parameterizations. We discretize the target curve into $M$ segments of uniform arc length $\Delta s$, which yields a spatial correspondence between the discrete unit index $j$ and the arc-length coordinate $s_j=j\,\Delta s$. This alignment allows us to compare the mechanism curvature $\kappa_j(\alpha_j,\phi_j(\Psi),l)$ directly against the target curvature sampled at the same location, $\kappa_j^t \equiv \kappa^t(s_j)$.

\subsubsection{Shape morphing as inverse design}

We pose the static shape-morphing task as an inverse-design problem: for a prescribed target curve, we seek mechanism parameters that minimize a shape-matching loss at a chosen actuation angle. Specifically, we treat the design variables as
\begin{equation}
\mathbf{x} = \{\,l,\;\Psi,\;\beta_0,\;\alpha_1,\dots,\alpha_M\,\}^{\mathsf{T}},
\end{equation}
and solve for $\mathbf{x}$ by minimizing a weighted least-squares objective that measures mismatch between the deployed mechanism and the target.

The loss is composed of $(i)$ a curvature-matching term evaluated at corresponding arc-length locations; $(ii)$ a tip-position term that enforces $\mathbf r_M\approx \mathbf p_M$; $(iii)$ an orientation-matching term that aligns the initial mechanism orientation $\beta_0$ with the initial tangent direction $\theta_{\mathrm{target}}$ of the target curve. The resulting objective is
\begin{align}
\mathcal{L}[\mathbf{x}] &=
\lambda_{\kappa} \sum_{j=1}^M\big(\kappa_j(\alpha_j,\phi_j,l)-\kappa_j^t\big)^2 \nonumber \\
&\quad + \lambda_{\tip} \|\r_M-\mathbf{p}_{\mathrm{M}}\|^2 \nonumber \\
&\quad + \lambda_{\mathrm{rot}} (\beta_0 - \theta_{\mathrm{target}})^2.
\end{align}

This optimization is implemented within a differentiable simulation framework, where exact gradients of the error/loss function with respect to the decision vector, $\nabla_{\mathbf{x}} \mathcal{L}$, are efficiently computed via automatic differentiation. These gradients are subsequently supplied to an adaptive, gradient-based optimizer to iteratively update the parameters until the mechanism converges to the target shape.

\begin{figure*}[t]
    \centering
    \includegraphics[width=\textwidth]{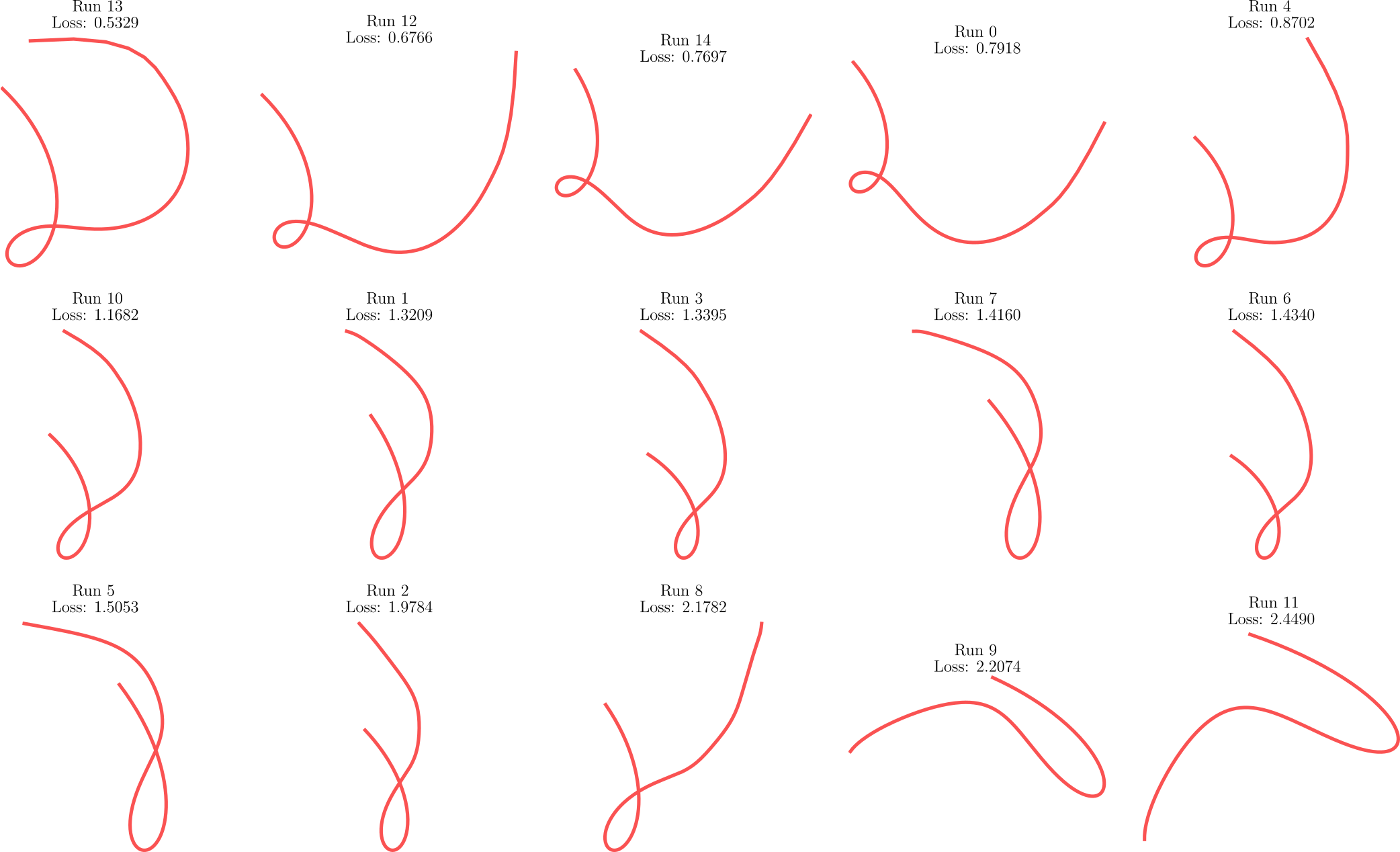}
    \caption{Collage of the tip trajectories for a mechanism with $N=50$ and 15 independent realizations using random initializations. The minimum (Run 13, Loss: 0.5329) is selected as the optimal configuration, while other runs capture varying degrees of target character fidelity or converge to suboptimal local minima.}
    \label{fig:SIfig4}
\end{figure*}

\section{Implementation of Trajectory Tracing via Curvature Matching}
\label{DynamicOptDetails}

The inverse design problem for trajectory tracing is formulated as a fully differentiable computational pipeline. The objective is to identify mechanism design parameters that reproduce a prescribed planar trajectory by matching its intrinsic curvature profile. The pipeline converts raw target coordinate data into an arc-length parameterized curvature representation, simulates the mechanism actuation through a forward kinematic model, and optimizes the design parameters using gradient-based methods enabled by automatic differentiation. The complete algorithm is illustrated schematically in Fig.~\ref{fig:SIfig3}.

\begin{figure*}[t]
    \centering
    \includegraphics[width=\textwidth]{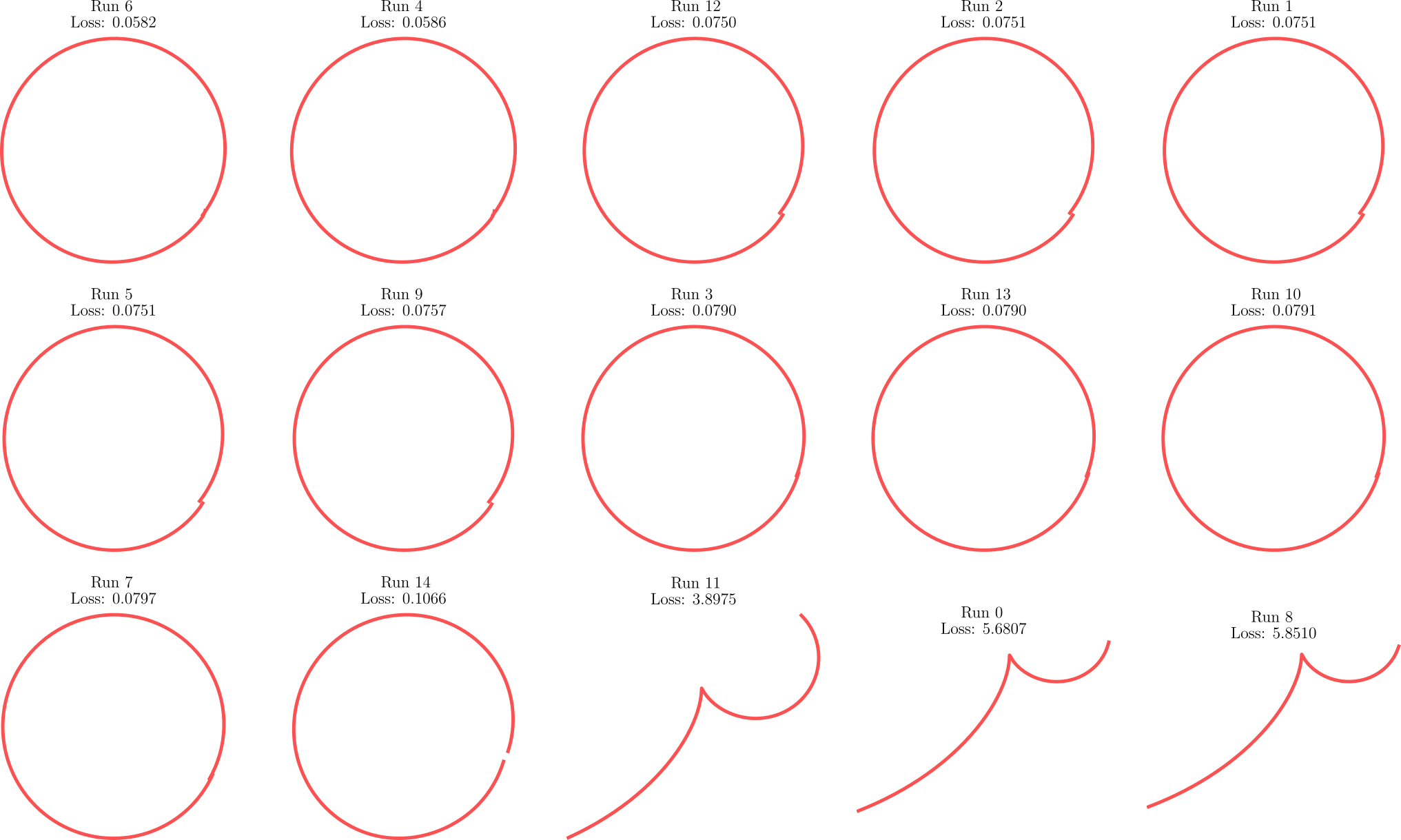}
    \caption{Collage of the tip trajectories from 15 independent optimization runs for a fixed number of units $N = 8$. While many runs converge to the target circle with low loss (e.g., Run 6, Loss: 0.0582), some realizations fall into local minima with high loss (e.g., Run 8, Loss: 5.8510), highlighting the complex optimization landscape.}\label{fig:SIfig5}
\end{figure*}

\subsection{Target shape processing and extracting curvature}
The target trajectory is provided as an ordered set of planar coordinates $\{(x_i, y_i)\}_{i=1}^N$ (e.g., obtained by image tracing). The coordinates are first normalized to a unit bounding box while preserving aspect ratio. An arc-length parameterization and the associated curvature profile are then constructed as in Sec.~\ref{StaticDesignDetails}. We denote the resulting smoothed target curvature by $\kappa^t(s)$, sampled on a uniform grid in $s\in[0,1]$.

\subsection{Differentiable forward model and arc-length reparameterization}

The mechanism geometry is naturally parameterized by an actuation variable $\Psi$, whereas the target curvature is defined with respect to arc length. A differentiable reparameterization is therefore required to enable meaningful comparison between the two.

\paragraph{Forward kinematics.}
Given a set of design parameters $\boldsymbol{\theta} = \{\alpha_1,\dots,\alpha_M, l\}$, where $\alpha_j$ denotes the aspect ratio of section $j$ and $l$ is the length of a single arm in the mechanism, the forward kinematics generates the tip trajectory $\mathbf{r}_{\mathrm{tip}}(\Psi) = (x(\Psi), y(\Psi))$ over a discretized actuation angle grid $\boldsymbol{\Psi} = [\Psi_0, \dots, \Psi_K]$. Here, $M$ is the number of contiguous constant-aspect-ratio sections used to approximate the desired trajectory (see Sec.~\ref{sec:GridSearch} for the discrete-parameter selection), and the actuation is swept monotonically from $\Psi_{\max}$ to $\Psi_{\min}$ (i.e., $\Psi_0 = \Psi_{\max}$ and $\Psi_K = \Psi_{\min}$).

\paragraph{Curvature of the mechanism parameterized by arc-length.}
To obtain an intrinsic curvature description, the trajectory is first reparameterized by arc length. Here, the superscript $M$ indicates a quantity computed from the mechanism trajectory (to distinguish it from the target-curve arc-length grid). The cumulative arc length of the mechanism trajectory is computed as
\begin{equation}
    s^M_k(\boldsymbol{\theta}) =
    \sum_{j=1}^{k} \sqrt{\big(x(\Psi_j) - x(\Psi_{j-1})\big)^2 + \big(y(\Psi_j) - y(\Psi_{j-1})\big)^2},
\end{equation}
and normalized to $[0,1]$. Thus, for each $k$, $s_k^M(\boldsymbol{\theta})$ is a scalar (the normalized arc-length at sample $k$), and we write $s_k \equiv s_k^M(\boldsymbol{\theta})$ for brevity. This defines a mapping $\Psi \mapsto s(\Psi;\boldsymbol{\theta})$ and allows us to treat the tip trajectory as an intrinsic curve $\mathbf r_{\tip}(s)$; the corresponding curvature is evaluated using differentiable finite differences on the uniform $s$-grid,
\begin{equation}
    \kappa_{\tip}(s_k) =
    \frac{x'_k y''_k - y'_k x''_k}
    {\left((x'_k)^2 + (y'_k)^2\right)^{3/2}},
\end{equation}
where derivatives are taken with respect to $s$.

\paragraph{Arc-length alignment.}
As in Sec.~\ref{StaticDesignDetails}, we compare curvature profiles on a common arc-length coordinate. Here, however, the arc-length distribution $s^M(\boldsymbol{\theta})$ depends on the design parameters, so the discrete samples $\{s_k\}_{k=0}^K$ generally do not coincide with the fixed target arc-length grid $\{s_k^{t}\}_{k=0}^K$. To preserve a pointwise comparison during optimization, we interpolate the computed curvature values $\{\kappa_{\tip}(s_k)\}_{k=0}^K$ onto $\{s_k^{t}\}_{k=0}^K$ using a differentiable linear interpolation operator. We denote the resulting resampled (interpolated) curvature vector by $\tilde{\boldsymbol{\kappa}}_{\tip}(\boldsymbol{\theta}) = \{ \tilde{\kappa}(s^M_k) \}$, where the tilde indicates interpolation onto the fixed target grid, $s^M_k(\boldsymbol{\theta})$. This quantity is directly comparable to the target curvature vector $\boldsymbol{\kappa}^t = \{\kappa^t_j\}$, evaluated on the same grid.

\subsection{Gradient evaluation using automatic differentiation}
The entire pipeline for evaluating the gradient of the error is implemented in \texttt{PyTorch}, which allows gradients to be propagated through the forward kinematics, arc-length computation, curvature evaluation, and interpolation layers. The optimization variables $\{ \tilde{\alpha}_j \} \in \mathbb{R}^M$ and $\tilde{l} \in \mathbb{R}$ are defined in an unconstrained space, and are mapped to physically meaningful parameters using a smooth transformation,
\begin{equation}
\alpha_j = \alpha_{\min} + (\alpha_{\max} - \alpha_{\min}) \, \mathcal{H}(\tilde{\alpha}_j), \qquad l = \exp(\tilde{l}),
\end{equation}
where $\mathcal{H}(\circ)$ is the Heaviside function. With this map, the geometric sequence $\alpha_j$ is strictly bounded within $[\alpha_{\min}, \alpha_{\max}]$.

The error is defined as a composite objective containing curvature mismatch, smoothness regularization, arc-length consistency, and a steric penalty,
\begin{equation}
    \mathcal{L}[\boldsymbol{\theta}] =
    \left\| \tilde{\boldsymbol{\kappa}}_{\tip} - \boldsymbol{\kappa}^t \right\|^2
    + \lambda_{\mathrm{sm}} \left\| \nabla \boldsymbol{\alpha} \right\|_2^2
    + \lambda_{L} \left( \frac{L - L^t}{L^t} \right)^2
    + \lambda_{\phi} \sum_{k=0}^K \sum_{j=1}^M \Big(\max\big(0, \phi_{\min} - \phi_j(\Psi_k)\big)\Big)^2.
\end{equation}
Here, $\boldsymbol{\alpha} = \{\alpha_j\}$, $L$ and $L^t$ denote the total mechanism length and the target arc length, respectively. The final term enforces a soft feasibility constraint during the actuation sweep by penalizing configurations for which the internal angle $\phi_j(\Psi_k)$ falls below a prescribed minimum value $\phi_{\min}$; this helps avoid near-singular configurations and ensures smooth tip motion. Gradient-based optimization follows the same differentiable simulation framework described in Sec.~\ref{StaticDesignDetails}; we use Adam together with automatic differentiation to compute $\nabla_{\boldsymbol{\theta}} \mathcal{L}$ efficiently through the computational graph.

\section{Grid Search for Discrete Parameter Selection}
\label{sec:GridSearch}

While the differentiable pipeline described in Sec.~\ref{DynamicOptDetails} optimizes the continuous aspect ratios, the number of units $N$ serves as a discrete hyperparameter. In the provided study (e.g., for the ``$D$'' character trajectory), we discretize the search space for $N \in [10, 100]$ with a step size of 5. For each candidate $N$, we instantiate a mechanism with $N$ units and perform the inverse design optimization described above. However, because the trajectory optimization landscape is non-convex and prone to local minima, a single descent run is insufficient. Therefore, we execute 15 independent optimization runs for each $N$, initialized with random aspect ratio values(representative outcomes for circular and ``$D$'' character trajectories are shown in Fig.~\ref{fig:SIfig5} and Fig.~\ref{fig:SIfig4}, respectively). The final design is selected as the global minimum of the composite loss $\mathcal{L}[\boldsymbol{\theta}]$ across all realizations, ensuring the solution represents the mechanism's true geometric capability rather than an initialization artifact.

\begin{figure*}[t]
    \centering
    \includegraphics[width=0.5\textwidth]{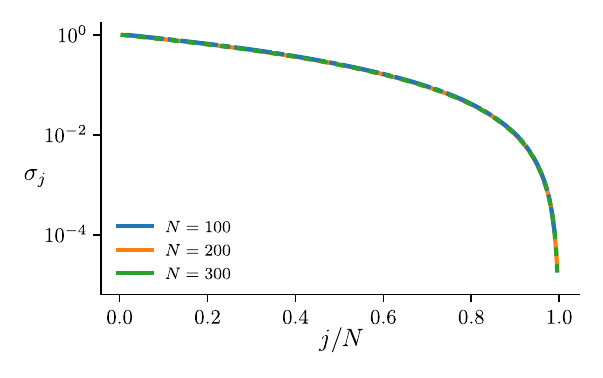}
    \caption{(A) The plot displays the sensitivity $\sigma_j$, defined as the variance of the tip displacement distribution $p(\Delta r^{(j)})$, against the normalized unit index $j/N$. Results for different mechanism sizes ($N = 100, 200, 300$) collapse onto a single curve, showing that the sensitivity profile is independent of the total number of units when scaled by the mechanism's length. The data indicates that the distal tip position is most sensitive to perturbations in units located near the base ($j/N \to 0$) and becomes progressively more robust to geometric errors as the unit index approaches the tip ($j/N \to 1$).}
    \label{fig:SIfig6}
\end{figure*}

\section{Sensitivity of tip position to changes in $\alpha_j$} \label{Sensitivity}
To understand how local variations in aspect ratio $\{\alpha_j\}$ influence the global deployed state, we performed a sensitivity analysis. We begin with a nominal configuration, $\balpha^0 = (\alpha_1^0, \dots, \alpha_N^0)$, corresponding to a straight deployment, i.e., $\alpha_j^0 = 0.5 \ \forall j$. At this value, the intrinsic curvature of each unit vanishes and the deployed configuration is straight. To isolate the influence of a single unit, we perturb only the $j$-th aspect ratio while keeping all other aspect ratios fixed. For each trial, we set $\alpha_j = \alpha_j^0 + \delta$, with $\delta$ sampled from a uniform distribution, $\delta \sim \mathcal{U}[-\epsilon, \epsilon]$. Now we define the perturbed parameter as $\balpha^{(j)} = \balpha^0 + \delta \, \mathbf{e}_j$, where $\mathbf{e}_j$ is the $j$-th Cartesian basis vector.
With $\r_\tip(\balpha)$ denoting the distal tip position obtained from the forward kinematics, the tip displacement induced by the perturbation is measured by
$\Delta r^{(j)} = \left\| \r_\tip(\balpha^{(j)}) - \r_\tip(\balpha^0) \right\|$. $\Delta r^{(j)}$ is obtained for multiple independent samples of $\delta$ resulting in a distribution, $p(\Delta r^{(j)})$. The sensitivity of the tip position to perturbations in the variance of tip displacement distribution $j$-th unit is then defined as the variance of this distribution,
$\sigma_j = \mathbb{E}\big[ (\Delta r^{(j)})^2\big] - \mathbb{E}\big[ (\Delta r^{(j)})\big]^2$. A larger $\sigma_j$ indicates that small perturbations in the $j$-th unit's geometry propagate strongly to the final tip position, whereas a smaller value implies the mechanism is locally robust to errors at that location. The log-linear trend in Fig.~\ref{fig:SIfig6} indicates that the tip position is exponentially sensitive to small perturbations in proximal units, while the sensitivity $\sigma_j$ drops by several orders of magnitude as the unit index approaches the distal end.

\section{Forward kinematics of segmented scissor mechanisms}\label{SI:segFK}
Consider a scissor mechanism composed of $N$ units with a uniform aspect ratio $\alpha$. Due to the symmetry of identical units, the mechanism's spatial configuration at any fixed actuation angle $\Psi$ forms a circular arc, characterized by an effective curvature $\kappa(\alpha,\Psi)$. While the mechanism's shape is circular for a fixed $\Psi$, the trajectory traced by the tip as $\Psi$ varies is generally non-circular because the curvature radius $\kappa^{-1}$ changes continuously with actuation.

We define the base configuration by the starting position $\mathbf{r}_0$ and the initial orientation vector $\hat{\mathbf{p}}_1$ (Fig.~\ref{fig:SIfig1}C). These are arbitrary choices that set the reference position and orientation for the forward-kinematics construction. The instantaneous center of curvature for this uniform assembly, denoted by $\tilde{\mathbf{q}}_1$, lies along this direction,
\begin{equation}
\tilde{\mathbf{q}}_1 = \mathbf r_0 - \kappa^{-1}(\alpha,\Psi)\,\hat{\mathbf p}_1 .
\end{equation}
The tip position is obtained by rotating the radius vector $\mathbf r_0 - \tilde{\mathbf{q}}_1$ using $\mathcal R^{\,N-1}(\phi_1^*)$,
\begin{equation}
\mathbf r_{\text{tip}}
=
\tilde{\mathbf{q}}_1
+
\mathcal R^{\,N-1}(\phi_1^*)\left(\mathbf r_0 - \tilde{\mathbf{q}}_1\right).
\end{equation}
Substituting for $\tilde{\mathbf{q}}_1$ and rearranging yields
\begin{equation}
\mathbf r_{\text{tip}}
=
\mathbf r_0
+
\kappa^{-1}(\alpha,\Psi)
\left(
\mathcal R^{\,N-1}(\phi_1^*) - \I
\right)\hat{\mathbf p}_1 .
\end{equation}
To trace more complex trajectories, we assemble the mechanism as $J$ contiguous sections, $j = 1,\dots,J$, where section $j$ contains $N_j$ units of constant aspect ratio $\alpha_j$ (see Fig.~\ref{fig:SIfig1}C for a schematic of a two-segment assembly). Each section is characterized by a constant $\phi_j^*(\alpha_j,\Psi)$ and an associated curvature $\kappa_j(\alpha_j,\Psi)$. To maintain geometric continuity, the center of curvature is shifted at the interface between adjacent sections, since the curvature (and hence the local radius and center of curvature) changes discontinuously when the aspect ratio changes from $\alpha_j$ to $\alpha_{j+1}$. The magnitude of this jump, denoted by $\zeta_j^{j+1}$ and illustrated as the shift between successive centers of curvature $\tilde{\mathbf{q}}_1$ and $\tilde{\mathbf{q}}_2$ in Fig.~\ref{fig:SIfig1}C, is derived from the unit geometry as
\begin{equation}
\zeta_j^{j+1}
=
\alpha_j \frac{\sin(\phi_{j}^*)}{\cos(\phi_{j}^*)}
-
\alpha_{j+1} \frac{\sin(\phi_{j+1}^*)}{\cos(\phi_{j+1}^*)}.
\end{equation}

Let $\tilde{\mathbf q}_j$ denote the center of curvature and $\hat{\mathbf p}_j$ the local orientation vector at the start of section $j$. These quantities evolve recursively as,
\begin{equation}
\tilde{\mathbf q}_{j+1}
=
\tilde{\mathbf q}_j
+
\zeta_j^{j+1}\,
\mathcal R(\Omega_j)\hat{\mathbf p}_j,
\qquad
\hat{\mathbf p}_{j+1}
=
\mathcal R(\Omega_j)\hat{\mathbf p}_j,
\end{equation}
where $\mathcal R(\cdot)$ is the planar rotation matrix. The rotation angle $\Omega_j$ for the first section, $\Omega_1 = {(2N_1 - 1)}\,\phi_1^*/2$, while for all subsequent intermediate sections, $\Omega_j = N_j\,\phi_j^*, j=2,\dots,J-1$. The tip of the mechanism lies at a half-unit offset from the start of the final section. Defining $\Omega_{\text{tip}} = {(2N_J - 1)}\phi_J^*/{2}$, the tip position is given by,
\begin{equation}
\mathbf r_{\text{tip}}(\Psi)
=
\tilde{\mathbf q}_J
+
\kappa_J^{-1}(\alpha_J,\Psi)\,
\mathcal R(\Omega_{\text{tip}})\hat{\mathbf p}_J .
\end{equation}
Iterating the recursive definitions and using
\begin{equation}
\hat{\mathbf p}_J
=
\left(
\prod_{m=1}^{J-1}\mathcal R(\Omega_m)
\right)\hat{\mathbf p}_1,
\end{equation}
the final center of curvature can be written explicitly as,
\begin{equation}
\tilde{\mathbf q}_J
=
\mathbf r_0
-
\kappa_1^{-1}(\alpha_1,\Psi)\hat{\mathbf p}_1
+
\sum_{k=1}^{J-1}
 \zeta_k^{k+1}
\left(
\prod_{m=1}^{k}\mathcal R(\Omega_m)
\right)\hat{\mathbf p}_1.
\end{equation}

Substituting this expression into the tip position and combining rotations yields the closed-form forward kinematics (which accurately describes the iterated section-wise rotations and translations shown in Fig.~\ref{fig:SIfig1}C),
\begin{equation}
\mathbf r_{\text{tip}}(\Psi)
=
\mathbf r_0
+
\left[
-\kappa_1^{-1}(\alpha_1,\Psi)\,\mathcal{I}
+
\sum_{k=1}^{J-1}
\zeta_k^{k+1}
\left(
\prod_{m=1}^{k}\mathcal R(\Omega_m)
\right)
+
\kappa_J^{-1}(\alpha_J,\Psi)
\left(
\prod_{m=1}^{J}\mathcal R(\Omega_m)
\right)
\right]\hat{\mathbf p}_1,
\end{equation}
where $\Omega_J \equiv \Omega_{\text{tip}}$.

\section{Experimental Setup}\label{ExpSetup}
\begin{figure*}[t]
    \centering
    \includegraphics[width=0.5\textwidth]{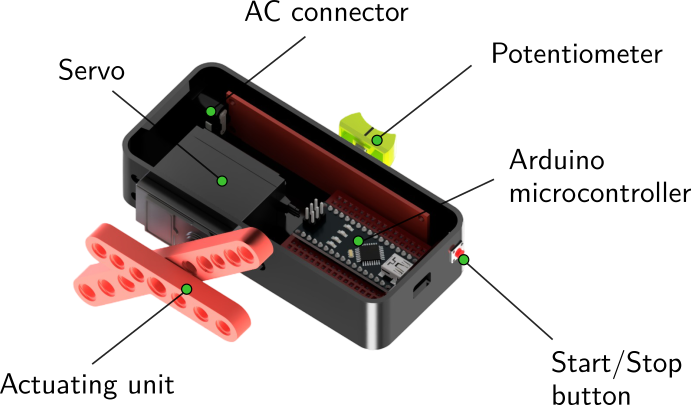}
    \caption{Three-dimensional view of CAD model of the controller-actuator setup. This setup includes a high-torque \textbf{servo} that drives the actuating unit; an \textbf{Arduino Nano microcontroller} processing logic; a \textbf{potentiometer} for manual angle control; a \textbf{start/stop button} for simple triggering; and an \textbf{AC connector} for stable external power.}
    \label{fig:SIfig7}
\end{figure*}

Our prototype of the controller-actuator setup to test the theoretical model is made using a \textbf{TowerPro MG995 servo} (stall torque 13 to 15 kg-cm) to actuate the single degree-of-freedom $\Psi$. As shown in Fig.~\ref{fig:SIfig7}, the servo is coupled to the \textbf{actuating unit} and controlled by an \textbf{Arduino microcontroller} housed in a custom 3D-printed enclosure. The control logic is straightforward; an analog \textbf{potentiometer} (in Fig.~\ref{fig:SIfig7}) maps user input directly to the servo angle, allowing us to manually deploy the mechanism and hold it at specific configurations for measurement. A \textbf{start/stop button} is included to trigger simple open-loop sweeping motions when needed. Stable power is supplied through an \textbf{AC connector} to ensure that the servo maintains sufficient torque against joint friction.
\end{document}